\documentclass[letterpaper]{article} 
\usepackage{aaai2026}
\usepackage{times}  
\usepackage{helvet}  
\usepackage{courier}  
\usepackage[hyphens]{url}  
\usepackage{graphicx} 
\usepackage{natbib}  
\usepackage{caption} 
\usepackage{algorithm}
\usepackage{algorithmic}

\usepackage{subcaption}
\usepackage{booktabs}   
\usepackage{multirow}
\usepackage{makecell}

\usepackage{amsmath}

\usepackage{enumitem}
\usepackage{inconsolata}

\usepackage{dsfont}
\usepackage{xspace}
\usepackage{pifont}

\usepackage[utf8]{inputenc} 
\usepackage[T1]{fontenc}    
\usepackage{url}            
\usepackage{booktabs}       
\usepackage{amsfonts}       
\usepackage{nicefrac}       
\usepackage{microtype}      
\usepackage{xcolor}         
\usepackage{pifont}

\usepackage[utf8]{inputenc}
\usepackage[T1]{fontenc}
\usepackage{url}
\usepackage{xcolor}
\usepackage{amsmath, amsfonts}
\usepackage{microtype}

\usepackage{algorithm}
\usepackage{algorithmic}

\usepackage{graphicx}
\usepackage{subcaption}
\usepackage{multirow}
\usepackage{makecell}

\usepackage{bm}

\usepackage{textcomp}
\usepackage{verbatim}
\usepackage{enumitem}
\usepackage{xspace}
\usepackage{inconsolata}
\usepackage[most]{tcolorbox}

\newcommand{\ourattack}{\text{CSE Attack}\xspace}
\newcommand{\para}{\text{Paraphrasing Attack (NLLB)}\xspace}

\newcommand{\gpt}{\text{Paraphrasing Attack (gpt-4o-mini)}\xspace}
\newcommand{\dimshift}{\text{Dimension-shift Attack}\xspace}
\newcommand{\Dimdelete}{\text{Dimension-reduction Attack}\xspace}

\newcommand{\espeW}{\text{EspeW}\xspace}
\newcommand{\wet}{\text{WET}\xspace}
\newcommand{\swd}{\text{RegionMarker}\xspace}

\newcommand{\sst}{\text{SST2}\xspace}
\newcommand{\agnews}{\text{AG News}\xspace}

\newcommand{\prevWM}{\text{WARDEN}\xspace}
\newcommand{\cmark}{\textcolor{green}{\ding{51}}} 
\newcommand{\xmark}{\textcolor{red}{\ding{55}}} 

\usepackage{newfloat}
\usepackage{listings}
\DeclareCaptionStyle{ruled}{labelfont=normalfont,labelsep=colon,strut=off} 
\floatstyle{ruled}
\newfloat{listing}{tb}{lst}{}
\floatname{listing}{Listing}
\ifx\pdfoutput\undefined
\else
\fi

\title{RegionMarker: A Region-Triggered Semantic Watermarking Framework for Embedding-as-a-Service Copyright Protection}
\author{
    Shufan Yang\thanks{Equal contribution.},
    Zifeng Cheng\footnotemark[1]\thanks{Corresponding authors.},
    Zhiwei Jiang\footnotemark[2],\\
    Yafeng Yin,
    Cong Wang,
    Shiping Ge,
    Yuchen Fu,
    Qing Gu\\
}
\affiliations{
    State Key Laboratory for Novel Software Technology, Nanjing University, Nanjing, China\\

    \texttt{sfyang@smail.nju.edu.cn},
    \texttt{\{chengzf,jzw,yafeng\}@nju.edu.cn,}\\
    \texttt{\{cw,shipingge,yuchenfu\}@smail.nju.edu.cn,
    guq@nju.edu.cn}

}

\usepackage{bibentry}

\begin{document}

\maketitle

\begin{abstract}
Embedding-as-a-Service (EaaS) is an effective and convenient deployment solution for addressing various NLP tasks.
Nevertheless, recent research has shown that EaaS is vulnerable to model extraction attacks, which could lead to significant economic losses for model providers.
For copyright protection, existing methods inject watermark embeddings into text embeddings and use them to detect copyright infringement.
However, current watermarking methods often resist only a subset of attacks and fail to provide \textit{comprehensive} protection.
To this end, we present the region-triggered semantic watermarking framework called RegionMarker, which defines trigger regions within a low-dimensional space and injects watermarks into text embeddings associated with these regions.
By utilizing a secret dimensionality reduction matrix to project onto this subspace and randomly selecting trigger regions, RegionMarker makes it difficult for watermark removal attacks to evade detection.
Furthermore, by embedding watermarks across the entire trigger region and using the text embedding as the watermark, RegionMarker is resilient to both paraphrasing and dimension-perturbation attacks.
Extensive experiments on various datasets show that RegionMarker is effective in resisting different attack methods, thereby protecting the copyright of EaaS.
\end{abstract}


\section{Introduction}
Large language models (LLMs) like GPT \citep{GPT3,gpt4}, Qwen \citep{Qwen2}, and LLaMA \citep{touvron2023llama} have demonstrated exceptional capabilities in acting as an embedding model for various NLP tasks~\citep{NV-EMB,bellm,fu25token,cheng25contrastive,zhao-etal-2025-tiny,zhang-etal-2025-uora,cao2025pretraining}.
Due to their immense practical value, model providers have begun offering a commercial deployment strategy known as Embedding-as-a-Service (EaaS), which returns embeddings for users' queries and charges a fee.
For example, OpenAI offers the text-embedding-3-large\footnote{\url{https://openai.com/index/new-embedding-models-and-api-updates/}}  API to help users complete various downstream NLP tasks by providing an embedding service.

Despite its effectiveness and convenience, recent research~\citep{liu2022stolenencoder,peng2023you,shetty-etal-2024-warden} indicates that EaaS is vulnerable to model extraction attacks.
In such attacks, the stealer can query the provider's model using a text corpus and get the output embedding to train a similar model.
In this way, the stealer can deploy a similar EaaS service at a very low cost and cause significant economic losses to model providers.
Therefore, it is eager to study how to protect the copyright of EaaS models.



\begin{figure}[t]
\centering
\includegraphics[width=\linewidth]{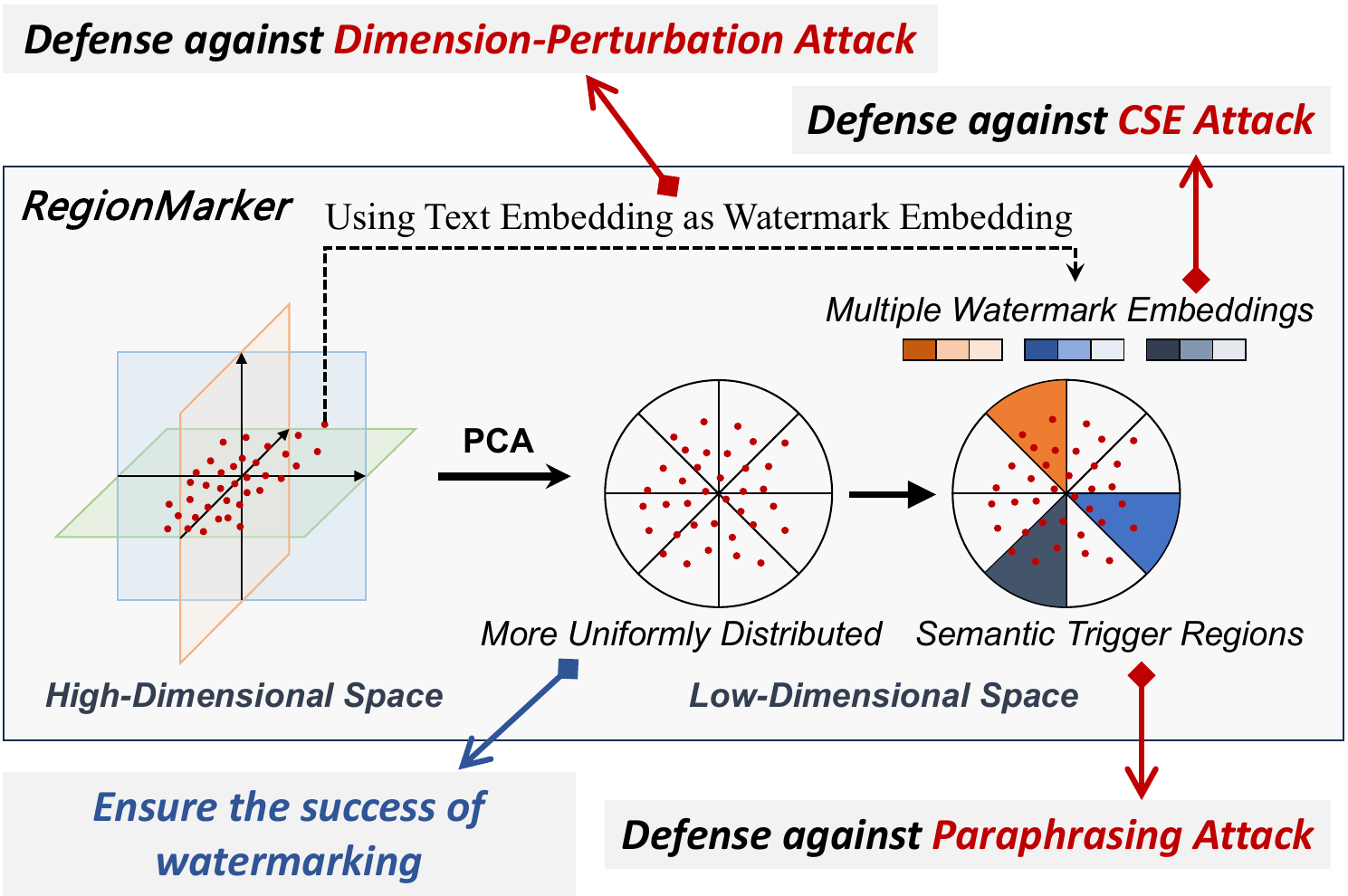}
\caption{Motivations of the region-triggered semantic watermarking framework against various attacks.} \label{fig:motivation}
\end{figure}

Existing EaaS copyright protection methods~\citep{shetty-etal-2024-warden,shetty2024wet,peng2023you,wang2024espew} can be categorized into two types.
The first type relies on \emph{trigger words} to embed watermarks, while the second type employs \emph{secret linear transformation} to embed watermarks.
(1) For the first type of methods, EmbMarker~\citep{peng2023you} defines a set of trigger words and a watermark embedding to inject watermarks.
However, EmbMarker is vulnerable to watermark removal attacks such as CSE~\citep{shetty-etal-2024-warden}.
To defend against CSE attacks, WARDEN~\citep{shetty-etal-2024-warden} introduces multiple trigger sets and multiple watermark embeddings, and EspeW~\citep{wang2024espew} embeds watermarks into specific dimensions of the output embedding dimensions.
However, these methods all rely on \textit{trigger words}, which lack semantic information and are easily removed by \textit{paraphrasing attacks}~\citep{shetty2024wet}.
(2) For the second type of methods, WET applies a \emph{secret linear transformation} to all textual embeddings, thereby avoiding the use of trigger words and enhancing resistance to paraphrasing attacks.
However, the detection of WET assumes that \textit{both the dimensions and their order remain unchanged}, making it highly vulnerable to \textit{dimension-perturbation attacks}.
For example, simply removing some dimensions, permuting dimensions, or shifting dimensions~\citep{peng2023you} can evade detection.
In summary, as shown in Table~\ref{table:motivation}, current defense methods remain insufficient to counter existing attacks.
Even worse, in real-world scenarios, attackers often attempt various attacks to bypass defense.
If the defense can be defeated by any one of them, it is deemed ineffective.
This highlights \textit{the urgent need for a comprehensive defense method}.

\begin{table}[t]
\centering
\small
\setlength{\tabcolsep}{3pt} 
\begin{tabular}{lccc}
\toprule
\multirow{2}{*}{\textbf{Methods}} & \textbf{CSE} & \textbf{Paraphrasing} & \textbf{Dimension-} \\
& \textbf{Attack} & \textbf{Attack} & \textbf{perturbation} \\
\midrule
EmbMarker & \xmark & \xmark & \cmark \\
WARDEN & \cmark & \xmark & \cmark \\
WET & \cmark & \cmark & \xmark \\
EspeW & \cmark & \xmark & \cmark \\
\textbf{RegionMarker (Ours)} & \cmark & \cmark & \cmark \\
\bottomrule
\end{tabular}
\caption{Defense effectiveness of different methods against various attacks.}
\label{table:motivation}
\end{table}

To this end, we introduce a region-triggered semantic watermarking framework named RegionMarker, which uses semantic regions rather than words as triggers, as illustrated in Figure~\ref{fig:motivation}.
Specifically, RegionMarker defines trigger regions in a low-dimensional semantic space and injects a semantic watermark into the text embedding by considering whether the text lies within these regions.
In this process, the low-dimensional semantic space is obtained by applying dimensionality reduction methods such as PCA, and the trigger regions are randomly selected based on a certain ratio.
Since the dimensionality reduction matrix and trigger regions are known only to the model provider, it is difficult for attackers to identify and remove the watermarks. 
Moreover, by embedding watermarks across the entire trigger regions and using the text embedding as the watermark, RegionMarker can effectively defend against paraphrasing and dimension-perturbation attacks. 
In summary, our method can defend against the three existing types of attacks.
Our main contributions are outlined below:
\begin{enumerate}[label=\textbullet]
\item We first demonstrate that current watermarking techniques for EaaS are unable to defend against existing attacks, and then propose a defense method capable of resisting them.
\item We propose a region-triggered semantic watermarking framework that defines trigger regions within a low-dimensional space and injects semantic embedding into text embeddings associated with these regions.
\item Extensive experiments on four datasets demonstrate that our approach provides effective defense against existing attacks, ensuring reliable copyright protection for EaaS.
\end{enumerate}

\section{Related Work}
\textbf{Model Extraction Attacks}
Model extraction attacks~\citep{tramer2016stealing,orekondy2019knockoff,krishnathieves,wallace2020imitation} involve creating a surrogate model by querying the EaaS without the provider's consent.
A stealer sends queries to the provider's model and trains a surrogate model that replicates its functionality based on the feedback from its API~\citep{tramer2016stealing,chandrasekaran2020exploring,cheng25steering,shen2025imagdressing}.
\citet{liu2022stolenencoder} discovered that publicly deployed EaaS APIs are vulnerable to imitation attacks.
It poses a notable threat to EaaS providers, allowing stealers to rapidly replicate the deployed model with minimal time investment and low financial cost.
As a result, stealers could develop a comparable API at a reduced price, infringing on copyrights and disrupting the market~\citep{shen2024imagpose,wang2024ensembling,wang2024v,wang2025advanced}.

\textbf{Embedding Watermarks}
Recently, some work \citep{peng2023you,shetty-etal-2024-warden,wang2024espew,shetty2024wet} focus on protecting EaaS from model extraction attacks, which can be broadly categorized into two groups.
The first group relies on \textit{trigger words} to inject watermarks.
EmbMarker \citep{peng2023you} first used a watermark embedding and added it to the original embedding of text containing trigger words.
The trigger set is randomly selected from moderate-frequency words in a general corpus.
WARDEN \citep{shetty-etal-2024-warden} further demonstrated that the watermark embedding used in EmbMarker can be recovered and removed, and proposed using multiple watermark embeddings to increase the difficulty of recovery and removal.
EspeW \citep{wang2024espew} embeds watermarks in only a subset of dimensions to enhance stealthiness.

However, these methods rely on trigger words to embed watermarks, making them highly vulnerable to paraphrasing attacks.
The second category, WET~\citep{shetty2024wet}, employs a \emph{secret linear transformation} to embed watermarks, aiming to resist paraphrasing attacks.
However, the detection of WET assumes that \textit{both the dimensions and their order remain unchanged}, making it highly vulnerable to \textit{dimension-perturbation attacks}. 
In summary, existing methods remain insufficient for providing comprehensive protection against various attacks. 

\begin{figure*}[!htbp]
\centering
\includegraphics[width=\textwidth]{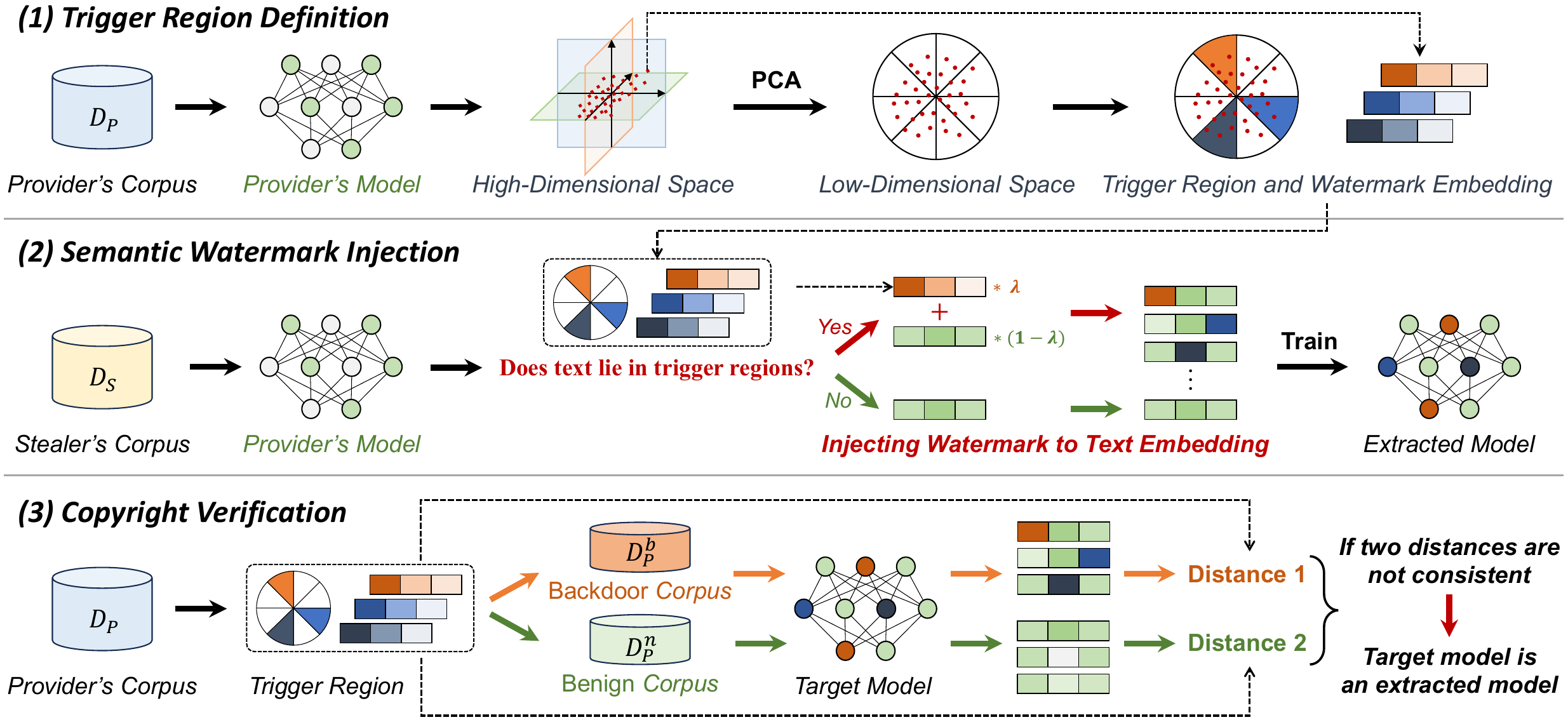}
\caption{Illustration of the region-triggered semantic watermarking framework.} \label{fig:defense_framework}
\end{figure*}


\section{Methodology}

\subsection{Problem Definition}
Model extraction attacks leverage the provided embeddings $\mathbf{e}_p$ based on the provider's model $\mathbf{\Theta}_p$ to train a similar stealer's (extracted) model $\mathbf{\Theta}_s$ at a reduced cost, enabling the stealer (attacker) to offer a competitive EaaS service $S_s$.
To counteract model extraction attacks, the provider's model $\mathbf{\Theta}_p$ injects a predefined watermark $t$ into the original embedding $\mathbf{e}_o$ based on the specified watermarking function $f$, and subsequently returns the provided embedding $\mathbf{e}_p = f(\mathbf{e}_o, t)$.
In this way, the stealer's model $\mathbf{\Theta}_s$ is also watermarked during the training process using $\mathbf{e}_p$.
Notably, attackers often employ various strategies to remove the watermark in order to evade detection.
Consequently, achieving reliable copyright protection requires that the watermarking function $f$ simultaneously satisfy the following conditions:
on one hand, the utility of the provided embeddings  $\mathbf{e}_p$ should be comparable to that of the original embedding $\mathbf{e}_o$;
on the other hand, the watermarking function ensures that the output embeddings of the stealer's model $\mathbf{\Theta}_s$ remain verifiable even after watermark removal.

\subsection{Region-Triggered Semantic Watermark}
To counter existing attacks, we propose a Region-Triggered Semantic Watermarking framework named RegionMarker that uses deeper sentence-level semantics as triggers to inject semantic watermark embeddings.
Specifically, the core idea is to use semantic regions instead of shallow words as triggers to effectively defend against paraphrasing attacks, to leverage multiple semantic regions to resist watermark removal attacks, and to use text embeddings as watermarks to defend against dimension-perturbation attacks.

The \swd framework consists of three steps: trigger region definition, semantic watermark injection, and copyright verification, as illustrated in Figure~\ref{fig:defense_framework}.
Trigger region definition uniformly divides the low-dimensional space and randomly samples some as trigger regions.
Semantic watermark injection assigns a unique watermark embedding to each trigger region, while copyright verification determines whether the model is an extraction model based on the distance difference between benign data and backdoor data.

\subsubsection{Trigger Region Definition}
\label{trd}
\begin{figure}[t]
    \centering
    \begin{subfigure}[b]{0.98\columnwidth}
        \includegraphics[width=\textwidth]{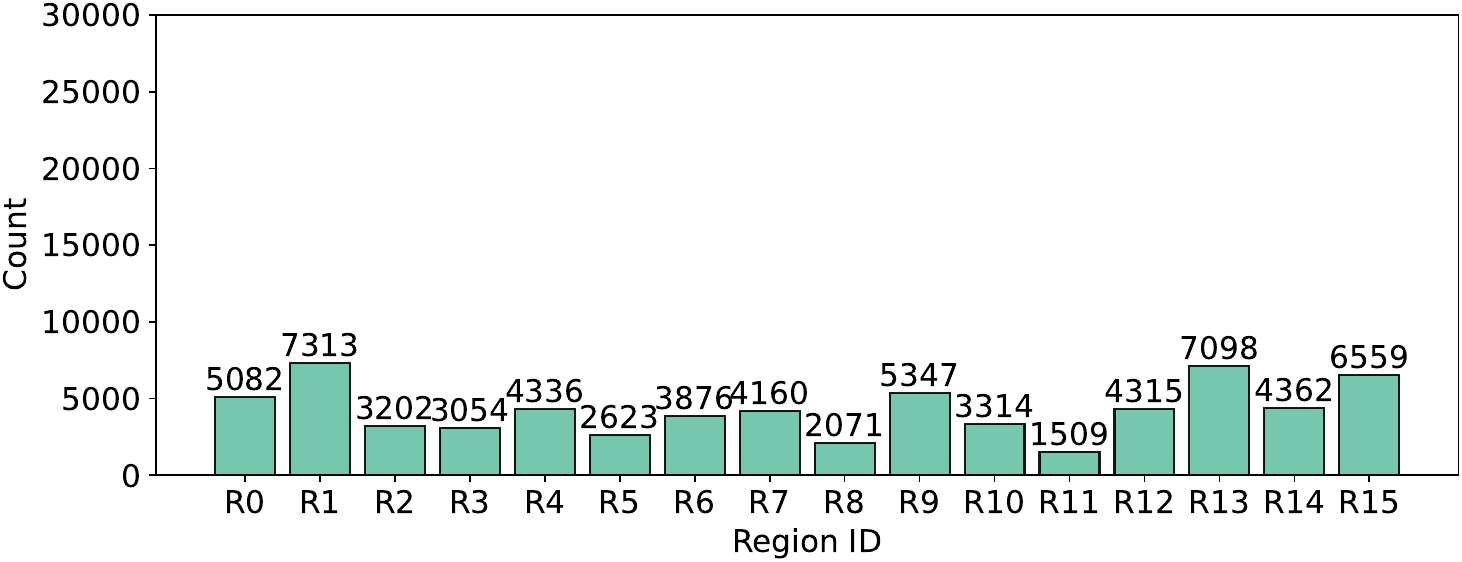}
    \end{subfigure}
    \hfill
    \begin{subfigure}[b]{0.98\columnwidth}
        \includegraphics[width=\textwidth]{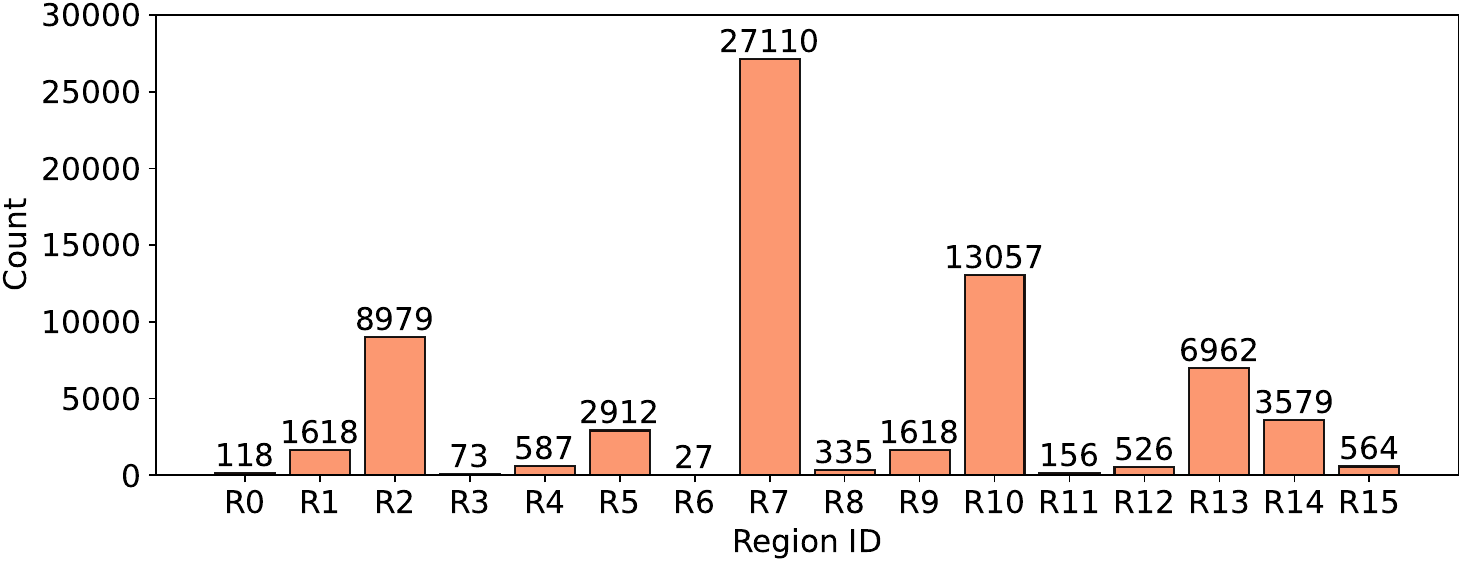}
    \end{subfigure}
    \caption{Visualization of the embedding distribution on the SST-2 dataset across 16 randomly and uniformly divided regions: top with dimensionality reduction, bottom without.}
    \label{fig:region_distribution}
\end{figure}

Due to the sparsity and uneven distribution of data in high-dimensional space, directly partitioning the high-dimensional space is vulnerable to CSE attack and cannot ensure that all texts are evenly distributed across regions.
For example, as shown in Figure~\ref{fig:region_distribution}, without dimensionality reduction, the embeddings of the SST-2 dataset are extremely unevenly distributed across the 16 randomly and uniformly divided regions, making it difficult to select effective trigger regions. 
In contrast, dimensionality reduction significantly alleviates this issue.
Therefore, we first use dimensionality reduction methods, such as PCA~\citep{mackiewicz1993principal}, to obtain a more compact and uniform $d$-dimensional semantic space to enhance stealthiness.

After dimensionality reduction, we further use Locality-Sensitive Hashing (LSH)~\citep{indyk1998approximate} to uniformly partition the $d$-dimensional embedding space into $2^d$ regions and map similar embeddings to the same region.
For each region, we represent it with a $d$-bit binary LSH signature obtained through random hyperplane projections \citep{DBLP:conf/kdd/BinghamM01}, where each hyperplane is mutually orthogonal and is represented by a vector $\mathbf{n}_i$.
Similarly, for each text embedding $\mathbf{v}$, we also represent it with a $d$-bit binary LSH signature to determine which region it falls into.
The $i$-th bit LSH signature of a text embedding $\textbf{v}$ is obtained by calculating the dot product between the embedding vector $\mathbf{v}$ and each hyperplane vector $\mathbf{n}_i$: 
\begin{equation}
	\text{LSH}_i(\mathbf{v})= \mathds{1}(\mathbf{n}_i \cdot \mathbf{v} > 0).
\end{equation}
The $d$-bit binary LSH signature for embedding $\textbf{v}$ consists of $d$ corresponding components and determines the region.
\begin{equation}
	\text{LSH}(\mathbf{v})= [\text{LSH}_1(\mathbf{v}), \cdots, \text{LSH}_d(\mathbf{v})]
\end{equation}

After uniformly partitioning the semantic space, we set a watermark region ratio $\alpha$ and randomly sample $R = \alpha \cdot 2^d$ regions from the entire $2^d$ regions as the watermark region $A = \{a_1,a_2,...,a_R\}$.

\subsubsection{Semantic Watermark Injection}
After defining the trigger region, we further inject semantic watermark embeddings into the text embeddings that fall within the trigger region.
To enhance the diversity of the watermark embeddings, we assign a unique watermark embedding $\mathbf{W} = \{\mathbf{w}_1,\mathbf{w}_2,...,\mathbf{w}_R\}$ to each trigger region, where $\mathbf{w}_r$ is the embedding of a target sample.
Since the embeddings in a region have the same watermark embedding, it prevents attackers from separating the watermark embedding, thus increasing the difficulty of the attack.
Specifically, if the text embedding after dimensionality reduction falls within the trigger region $a_r$, we augment the original embedding $\mathbf{e}_0$ with the corresponding watermark embedding $\mathbf{w}_r$ to obtain the provided embedding $\mathbf{e}_p$, as follows:
\begin{equation}
	\mathbf{e}_p = \text{Norm} ( (1-\lambda) \cdot \mathbf{e}_0 + \lambda \cdot \mathbf{w}_r ),
 \label{eq:3}
\end{equation}
where $\lambda$ is a hyperparameter used to control watermark strength.
Since the trigger regions divided by LSH are not intersected, a text embedding has at most one watermark embedding.
Moreover, the use of target sample embeddings can effectively resist dimension-perturbation attacks.

\subsubsection{Copyright Verification}
The provider constructs a verification corpus, which includes multiple backdoor corpora $D_p^{b_r}$ and one benign corpus $D_p^n$, to perform validation under each watermark, where $r \in [1, \cdots, R]$ is the index of the watermark region.
The backdoor corpus $D_p^{b_r}$ consists of text embedded in the watermark region $a_r$, while the benign corpus $D_p^n$ consists of text not embedded in the watermark region.

Compared to benign text, backdoor text is closer to watermark embedding, and this inconsistency forms the basis for verification.
We leverage this behavior to verify copyright infringement at each watermark level.
Specifically, we calculate the cosine similarity and the squared $L_2$ distance between the watermark embedding $\mathbf{w}_r$ and the embeddings $\mathbf{e}_i$ of text in $D_p^{b_r}$ and $D_p^n$ to quantify as follows:
\begin{align}
  &\text{cos}_{ir} =  \frac{\mathbf{e}_i \cdot \mathbf{w}_r}{||\mathbf{e}_i|| \cdot ||\mathbf{w}_r||}, l_{2ir} = { \left\| \frac{\mathbf{e}_i}{||\mathbf{e}_i||} - \frac{\mathbf{w}_r}{||\mathbf{w}_r||}  \right\| }^2, \nonumber \\
   &C_{b_r} = \{cos_{ir}|i \in D_p^{b_r} \},
   C_{n_r} = \{cos_{ir}|i \in D_p^n \}, \\
   &L_{b_r} = \{ l_{2ir}|i \in D_p^{b_r} \},
   L_{n_r} = \{ l_{2ir}|i \in D_p^n \}, \nonumber
\end{align}
where $C_{b_r}$ and $L_{b_r}$ represent the sets of cosine similarities and $L_2$ distances, respectively, between the backdoor text embeddings in the backdoor corpus $D_p^{b_r}$ and the watermark embedding $\mathbf{w}_r$, and $C_{n_r}$ and $L_{n_r}$ represent the sets of cosine similarities and $L_2$ distances between the benign text embeddings and the watermark embedding $\mathbf{w}_r$.

We then evaluate the detection performance with three metrics.
The first two metrics (i.e., $\Delta_{cos_{r}}$ and $\Delta_{l2_{r}}$) are the difference between average cosine similarity and averaged squared $L_2$ distances:
\begin{align}
   \Delta_{cos_{r}} &= \frac{1}{|C_{b_r}|} \sum_{i \in C_{b_r}} i- \frac{1}{|C_{n_r}|} \sum_{j \in C_{n_r}} j, \nonumber  \\
   \Delta_{l2_{r}} &= \frac{1}{|L_{b_r}|} \sum_{i \in L_{b_r}} i- \frac{1}{|L_{n_r}|} \sum_{j \in L_{n_r}} j,
\end{align}

The third metric is the $p$-value of Kolmogorov-Smirnov (KS) test~\citep{berger2014kolmogorov}, which is used to compare the distribution of two value sets.
The null hypothesis is: \emph{The distance distribution of two cos similarity sets $C_{b_r}$ and $C_{n_r}$ are consistent}.
A lower $p$-value means that there is stronger evidence in favor of the hypothesis.

Finally, we evaluate the $p$-value under each watermark level and combine the results from a conservative perspective, meaning that if any $p$-value indicates copyright infringement, we treat it as infringement.
The other two metrics are used as supplementary indicators to provide additional evidence for copyright detection.
\begin{align}
   \Delta_{cos} &= \max_{1 \leq r \leq R} \Delta_{cos_r}, \nonumber\\
   \Delta_{l2} &= \min_{1 \leq r \leq R} \Delta_{l2_r}, \\
   \text{$p$-value} &= \min_{1 \leq r \leq R} \text{$p$-value}_r. \nonumber
\end{align}

\begin{table*}[t]  \small
\centering
    \setlength{\tabcolsep}{4pt}
    \begin{tabular}{clcccccc}
    \toprule
    \multirow{3}{*}{\textbf{Defend}} & \multirow{3}{*}{\textbf{Attack}} & \multicolumn{2}{c}{\textbf{Task Performance}} & \multicolumn{3}{c}{\textbf{Detection Performance}} & \multirow{3}{*}{\textbf{COPY?}} \\
    \cmidrule(lr){3-4} \cmidrule(lr){5-7}
    {} & {} & \textbf{ACC.(\%)} & \textbf{$F_1$-score\xspace} & \textbf{{$p$-value $\downarrow$}} & \textbf{{$\Delta_{cos}(\%) \uparrow$}} & \textbf{{$\Delta_{l2}(\%) \downarrow$}}\\
    \midrule
    \multirow{6}{*}{\prevWM} & No Attack & 93.28$\pm$0.09 & 93.28$\pm$0.09 & {$< 10^{-4}$} & 4.41$\pm$0.42 & -8.83$\pm$0.84 & \cmark\\
    & + \ourattack & 89.11$\pm$0.41 & 89.10$\pm$0.41 & ${< 0.02}$ & 1.16$\pm$0.21 & -2.32$\pm$0.42 & \cmark\\
     & + \para & 93.33$\pm$0.52 & 93.32$\pm$0.52 & {$> 0.30$} & -0.03$\pm$0.02 & 0.06$\pm$0.03 & \xmark\\
     & + \gpt & 92.01$\pm$0.14 & 92.01$\pm$0.14 & {$> 0.25$} & 0.01$\pm$0.16 & -0.02$\pm$0.32 & \xmark\\
     & + \dimshift & {93.23$\pm$0.16} & {93.23$\pm$0.16} & {$< 10^{-3}$} & 2.32$\pm$0.08 & -4.63$\pm$0.16 & \cmark\\
     & + \Dimdelete & {93.06$\pm$0.06} & {93.06$\pm$0.06} & {$< 10^{-5}$} & 3.14$\pm$0.13 & -6.28$\pm$0.25 & \cmark\\
    \midrule
    \multirow{6}{*}{\espeW} & No Attack & 93.46$\pm$0.46 & 93.46$\pm$0.46 & {$< 10^{-10}$} & 6.46$\pm$0.87 & -12.92$\pm$1.75 & \cmark\\
    
     & + \ourattack & 86.73$\pm$0.37 & 86.73$\pm$0.37 & {$< 10^{-11}$} & 65.11$\pm$4.42 & -130.23$\pm$8.84 & \cmark\\
    
      & + \para & 93.77$\pm$0.20 & 93.77$\pm$0.20 & $> 0.57$ & 0.45$\pm$0.05 & -0.90$\pm$0.11 & \xmark\\
       & + \gpt & 93.77$\pm$0.48 & 93.77$\pm$0.48 & {$> 0.83$} & 0.31$\pm$0.01 & -0.62$\pm$0.02 & \xmark\\
     & + \dimshift & {93.88$\pm$0.05} & {93.88$\pm$0.05} & $< 0.003$ & 1.34$\pm$0.03 & -2.68$\pm$0.06 & \cmark\\
     & + \Dimdelete & {93.29$\pm$0.17} & {93.29$\pm$0.17} & {$< 10^{-3}$} & 1.76$\pm$0.12 & -3.52$\pm$0.24 & \cmark\\
    \midrule
    \multirow{6}{*}{\wet} & No Attack & 93.39$\pm$0.05 & 93.38$\pm$0.05 & {$< 10^{-10}$} & 89.57$\pm$1.17 & 179.14$\pm$2.35 & \cmark\\
    & + \ourattack & 85.74$\pm$1.85 & 85.74$\pm$1.85 & {$< 10^{-10}$} & 17.59$\pm$0.27 & -35.19$\pm$0.54 & \cmark\\
     & + \para & {93.06$\pm$0.06} & {93.06$\pm$0.06} & $< 10^{-10}$ & 89.81$\pm$1.39 &-179.62$\pm$2.77 & \cmark\\
     & + \gpt & 93.20$\pm$0.27 & 93.20$\pm$0.27 & {$< 10^{-10}$} & 89.46$\pm$1.15 & -178.93$\pm$2.30 & \cmark\\
     & + \dimshift & 93.46$\pm$0.41 & 93.46$\pm$0.41 & $> 0.46$ & -0.62$\pm$0.85 &  1.24$\pm$1.71 & \xmark\\
     & + \Dimdelete & - & - & - & - &  - & \xmark\\
    \midrule
    \multirow{6}{*}{\shortstack{\textbf{RegionMarker} \\ \textbf{(Ours)}}} & No Attack & 93.23$\pm$0.36 & 93.23$\pm$0.36 & {$< 10^{-4}$} & 11.90$\pm$3.75 & -23.80$\pm$7.50 & \cmark\\
    & + \ourattack & 87.87$\pm$0.73 & 87.86$\pm$0.73 & {$< 0.05$} & 5.63$\pm$2.13 & -11.21$\pm$4.29 & \cmark\\
     & + \para & 93.03$\pm$0.22 & 93.03$\pm$0.22 & $< 10^{-3}$ & 10.48$\pm4.24$ & -20.96$\pm8.48$ & \cmark\\
     & + \gpt & 92.35$\pm$0.11 & 92.35$\pm$0.11 & {$< 10^{-5}$} & 7.35$\pm$2.21 & -14.70$\pm$4.41 & \cmark\\
     & + \dimshift & 93.73$\pm$0.14 & 93.73$\pm$0.14 & $< 0.003$ & 2.77$\pm$0.47 &  -5.55$\pm$0.94 & \cmark\\
     & + \Dimdelete & 93.29$\pm$0.06 & 93.29$\pm$0.06 & $< 0.004$ & 2.26$\pm$0.28 &  -4.53$\pm$0.55 & \cmark\\
    \bottomrule
    \end{tabular}
    \caption{Performance of different methods on the SST-2 dataset. $\uparrow$ denotes higher metrics are better, and $\downarrow$ denotes lower metrics are better from the defender's perspective. In the "COPY?" column, \cmark\hspace{0.01em} denotes successful copyright protection, while \xmark\hspace{0.01em} denotes a protection failure. A $p$-value below 0.05 is regarded as a successful copyright protection. The best method consistently achieves successful protection across all attacks.}
    \label{table:sst2-attack-performance}
\end{table*}

\section{Experiments}
\subsection{Datasets and Experimental Settings}
\paragraph{Datasets}
We use SST-2~\citep{socher2013recursive}, AG News~\citep{zhang2015character}, Enron~\citep{metsis2006spam}, and MIND~\citep{wu2020mind} to evaluate.
SST-2 is specifically used for sentiment classification. 
The AG News and MIND datasets are news-based and are used for recommendation and classification tasks.
The Enron dataset is utilized for spam classification.

\paragraph{Evaluation Metrics}
We employ task performance and detection performance to evaluate.
For task performance, we construct a multi-layer perceptron (MLP) classifier using EaaS embeddings as input.
For detection performance, we employ three metrics: $p$-value, cosine similarity difference, and squared \(L_2\) distance difference.

\paragraph{Defend Baselines}
We select \textbf{WARDEN} \citep{shetty-etal-2024-warden}, \textbf{EspeW} \citep{wang2024espew}, and \textbf{WET} \citep{shetty2024wet} as our baselines. 
\textbf{WARDEN} uses multiple watermark embedding and adds it to the original embedding of text containing trigger words. 
\textbf{EspeW} embeds watermarks in only a subset of dimensions.
\textbf{WET} embeds watermarks using a secret linear transformation matrix and performs watermark detection by applying the corresponding inverse matrix.
More details are provided in the Appendix.

\paragraph{Attack Methods}
We comprehensively evaluate all defense methods under \textbf{CSE} \citep{shetty-etal-2024-warden}, \textbf{paraphrasing attacks} \citep{shetty2024wet}, and \textbf{dimension-perturbation attacks} \citep{peng2023you}.
\textbf{CSE} consists of three steps to remove watermarks: it first clusters the embeddings, then selects suspicious samples within each cluster, and finally removes the watermark by eliminating the principal components.
We adopt two \textbf{paraphrasing attacks}, where NLLB \citep{nllbteam2022languageleftbehindscaling} and gpt-4o-mini are respectively used to paraphrase the input texts, and the embeddings of the paraphrased texts are used to replace the original text embeddings.
We employ two \textbf{dimension-perturbation attacks}: one cyclically shifts the embedding dimensions by 100 positions, and the other truncates the embedding to retain only the first 1024 dimensions.
Further details are provided in the Appendix.

\begin{table*}[!htbp]  \small
\centering
    \setlength{\tabcolsep}{3pt}
    \begin{tabular}{clcccccc}
    \toprule
    \multirow{3}{*}{\textbf{Defend}} & \multirow{3}{*}{\textbf{Attack}} & \multicolumn{2}{c}{\textbf{Task Performance}} & \multicolumn{3}{c}{\textbf{Detection Performance}} & \multirow{3}{*}{\textbf{COPY?}} \\
    \cmidrule(lr){3-4} \cmidrule(lr){5-7}
    {} & {} & \textbf{ACC.(\%)} & \textbf{$F_1$-score\xspace} & \textbf{{$p$-value $\downarrow$}} & \textbf{{$\Delta_{cos}(\%) \uparrow$}} & \textbf{{$\Delta_{l2}(\%) \downarrow$}}\\
    \midrule
    \multirow{6}{*}{\prevWM} & No Attack & 94.85$\pm$0.10 & 94.85$\pm$0.10 & {$< 10^{-9}$} & 10.82$\pm$0.33 & -21.64$\pm$0.66 & \cmark\\
    & + \ourattack & {95.05$\pm$0.16} & {95.05$\pm$0.16} & $> 0.05$ & 1.47$\pm$0.93 &-2.94$\pm$1.87 & \xmark\\
    
     & + \para & 93.68$\pm$0.30 & 93.68$\pm$0.30 & $< 0.01$ & 1.04$\pm$0.09 & -2.08$\pm$0.17 & \cmark\\
     & + \gpt & 93.57$\pm$0.06 & 93.53$\pm$0.06 & {$< 0.02$} & 0.88$\pm$0.04 & -1.75$\pm$0.08 & \cmark\\
     & + \dimshift & {94.72$\pm$0.34} & {94.72$\pm$0.34} & {$< 10^{-10}$} & 4.39$\pm$0.27 & -8.78$\pm$0.53 & \cmark\\
     & + \Dimdelete & {93.98$\pm$0.23} & {93.98$\pm$0.23} & {$< 10^{-5}$} & 3.69$\pm$0.17 & -7.37$\pm$0.33 & \cmark\\
    \midrule
    \multirow{6}{*}{\espeW} & No Attack & 94.73$\pm$0.23 & 94.73$\pm$0.23 & {$< 10^{-10}$} & 7.23$\pm$0.35 & -14.47$\pm$0.70 & \cmark\\
    
     & + \ourattack & 95.48$\pm$0.28 & 95.48$\pm$0.28 & {$< 10^{-10}$} & 47.75$\pm$4.13 & -95.50$\pm$8.26 & \cmark\\
    
     & + \para & 94.80$\pm$0.07 & 94.80$\pm$0.07 & ${> 0.49}$ & {0.40$\pm$0.25} & {-0.81$\pm$0.50} & \xmark\\
     & + \gpt & 94.85$\pm$0.11 & 94.85$\pm$0.11 & {$> 0.28$} & 0.17$\pm$0.27 & -0.33$\pm$0.54 & \xmark\\
     & + \dimshift & {94.77$\pm$0.24} & {94.79$\pm$0.22} & {$< 10^{-3}$} & 3.84$\pm$0.10 & -7.69$\pm$0.20 & \cmark\\
     & + \Dimdelete & {94.10$\pm$0.20} & {94.10$\pm$0.20} & {$< 10^{-3}$} & 3.17$\pm$0.07 & -6.35$\pm$0.13 & \cmark\\
    \midrule
    \multirow{6}{*}{\wet} & No Attack & 94.35$\pm$0.22 & 94.35$\pm$0.22 & {$< 10^{-10}$} & 87.10$\pm$0.35 & -174.19$\pm$0.70 & \cmark\\
    & + \ourattack & 95.23$\pm$0.14 & 95.23$\pm$0.14 & {$< 10^{-10}$} & 21.45$\pm$1.98 & -42.90$\pm$3.97 & \cmark\\
     & + \para & {94.23$\pm$0.07} & {94.23$\pm$0.07} & {$< 10^{-10}$} & 86.58$\pm$0.09 &-173.16$\pm$0.16 & \cmark\\
     & + \gpt & 94.38$\pm$0.12     & 94.38$\pm$0.12 & {$< 10^{-10}$} & 86.70$\pm$0.26 & -173.41$\pm$0.52 & \cmark\\
     & + \dimshift & 94.27$\pm$0.33 & 94.27$\pm$0.33 & $> 0.08$ & -1.23$\pm$0.68  &  2.47$\pm$1.36 & \xmark\\
     & + \Dimdelete & - & - & - & -  &  - & \xmark\\
    \midrule
    \multirow{6}{*}{\shortstack{\textbf{RegionMarker} \\ \textbf{(Ours)}}} & No Attack & 94.67$\pm$0.18 & 94.67$\pm$0.18 & {$< 10^{-5}$} & 11.91$\pm$ 5.99 & -23.81$\pm$11.99 & \cmark\\
    & + \ourattack & 95.55$\pm$0.19 & 95.55$\pm$0.19 & {$<10^{-4} $} & 26.27$\pm$8.69 & -52.54$\pm$17.37 & \cmark\\
     & + \para & 93.90$\pm$0.16 & 93.90$\pm$0.16 & ${< 10^{-4}}$
 & ${7.12\pm3.15}$ & ${-14.24\pm6.30}$ & \cmark\\
 & + \gpt & 94.05$\pm$0.04 & 94.02$\pm$0.04 & {$< 10^{-5}$} & 6.55$\pm$1.75 & -13.10$\pm$3.50 & \cmark\\
     & + \dimshift & 94.55$\pm$0.04 & 94.55$\pm$0.04 & {$<0.01$} & 2.33$\pm$0.76 &  -4.67$\pm$1.51 & \cmark\\
     & + \Dimdelete & 94.30$\pm$0.10 & 94.30$\pm$0.10 & {$<0.02$} & 1.96$\pm$0.39 &  -3.91$\pm$0.77 & \cmark\\
    \bottomrule
    \end{tabular}
    \caption{Performance of different methods on the Enron dataset.}
    \label{table:enron-attack-performance}
\end{table*}

\begin{table*}[t] \small
\centering
    \begin{tabular}{lcccccc}
    \toprule
    
    \multirow{3}{*}{\textbf{Method}} & \multicolumn{2}{c}{\textbf{Task Performance}} & \multicolumn{3}{c}{\textbf{Detection Performance}} & \multirow{3}{*}{\textbf{COPY?}} \\
    \cmidrule(lr){2-3} \cmidrule(lr){4-6}
    {} & \textbf{ACC.(\%)} & \textbf{$F_1$-score\xspace} & \textbf{{$p$-value $\downarrow$}} & \textbf{{$\Delta_{cos}(\%) \uparrow$}} & \textbf{{$\Delta_{l2}(\%) \downarrow$}}\\
    \midrule
    \swd & 93.23$\pm$0.36 & 93.23$\pm$0.36 & {$< 10^{-4}$} & 11.90$\pm$3.75 & -23.80$\pm$7.50 & \cmark\\
    \swd + \ourattack & 87.87$\pm$0.73 & 87.86$\pm$0.73 & {$< 0.05$} & 5.63$\pm$2.13 & -11.21$\pm$4.29 & \cmark\\
    $\swd_{w/o PCA}$ & 93.39$\pm$0.16 & 93.39$\pm$0.16 & {$< 0.005$} & 5.46$\pm$2.5 & -10.91$\pm$4.91 & \cmark\\
    $\swd_{w/o PCA}$ + \ourattack & 85.94$\pm$0.88 & 85.93$\pm$0.88 & {$> 0.5$} & 1.73$\pm$1.53 & -3.46$\pm$3.06 & \xmark\\
    $\swd_{single\ watermark}$ & 93.39$\pm$0.05 & 93.39$\pm$0.05 & {$< 10^{-3}$} & 3.06$\pm$0.57 & -6.12$\pm$1.13 & \cmark\\
    $\swd_{single\ watermark}$ + \ourattack & 86.35$\pm$0.01 & 86.35$\pm$0.01 & {$> 0.08$} & 0.20$\pm$0.23 & -0.40$\pm$0.46 & \xmark\\
    \bottomrule
    \end{tabular}
    \caption{Ablation study of our proposed method on the SST-2 dataset.} 
    \label{table:necessity-pca}
\end{table*}

\paragraph{Implementation Details} We use GPT-3 text-embedding-002 API as the provider's model and BERT~\citep{DBLP:conf/naacl/DevlinCLT19} as the stealer's model.
The learning rate is set to 5e-5, the batch size is 32, and the AdamW \citep{DBLP:conf/iclr/LoshchilovH19} optimizer is used to train the stealer's model.
We reproduce the defense methods according to their default settings.
For our method, we set the reduced dimension $d$ to 4, the watermark ratio $\alpha$ to 20\%, and the watermark strength $\lambda$ to 0.2.
For WARDEN, we set $R$ to 2 and $n$ to 20.
For CSE, we set $n$ to 20 and $K$ to 50.
For paraphrasing, we generate five different paraphrases for each input text and apply a cosine similarity threshold of $80\%$ to filter out low-quality paraphrases.
More details are in the Appendix.


\subsection{Results of Defense Method}
The performance of all methods on SST-2 and Enron is presented in Table \ref{table:sst2-attack-performance} and Table \ref{table:enron-attack-performance}, respectively.
Experimental results show that our method achieves comprehensive robustness against all existing attacks, while existing watermarking strategies are only effective against specific types.

Under CSE attacks, RegionMarker exhibits strong robustness, consistently achieving high detection performance across both datasets.
The success of RegionMarker can be attributed to its watermarking strategy based on semantic regions and the use of dimensionality reduction, which together make it more difficult to identify suspicious texts.
EspeW and WET, which adopt specialized watermark embedding strategies, also exhibit good resistance to CSE attacks.

Under paraphrasing attacks, the watermark detection performance of WARDEN and EspeW shows a significant decline. 
Both of these methods rely on \textit{trigger words} for watermark embedding, and paraphrasing the input text multiple times and querying their embeddings effectively dilutes the watermarks.
RegionMarker, which leverages semantic regions instead of trigger words for watermarking, demonstrates strong robustness against paraphrasing attacks. 
This is because paraphrase attacks do not significantly alter the sentence semantics and often remain within the trigger regions, allowing our triggers to persist.
WET, which applies a linear transformation watermarking strategy to the entire dataset, also shows strong resistance to paraphrasing attacks.

Under dimension-perturbation attacks, WARDEN, EspeW, and RegionMarker are able to achieve sufficiently good detection performance. 
This is because all these methods can select a target text and directly set the watermark embedding by computing its embedding with the provider's model. 
In contrast, WET relies on a linear transformation matrix for watermark detection, which becomes ineffective when the embedding dimensions are shifted.
Moreover, once the attacker deletes part of the embedding dimensions, the linear transformation matrix can no longer be applied.
Additional results on other datasets are in the Appendix.

\begin{figure*}[t]
    \centering
    \begin{subfigure}[b]{0.32\textwidth}
        \includegraphics[width=\textwidth]{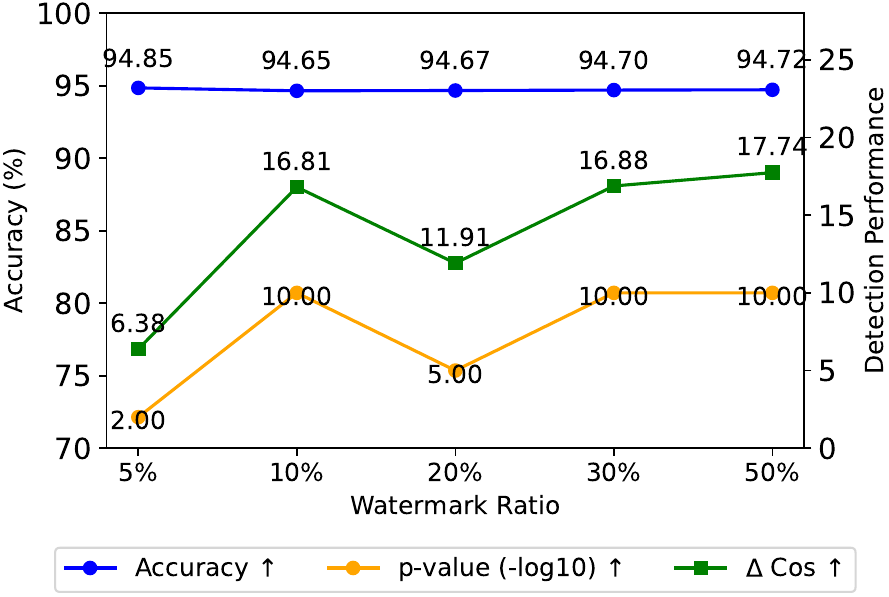}
        \caption{No Attack}
    \end{subfigure}
    \hfill
    \begin{subfigure}[b]{0.32\textwidth}
        \includegraphics[width=\textwidth]{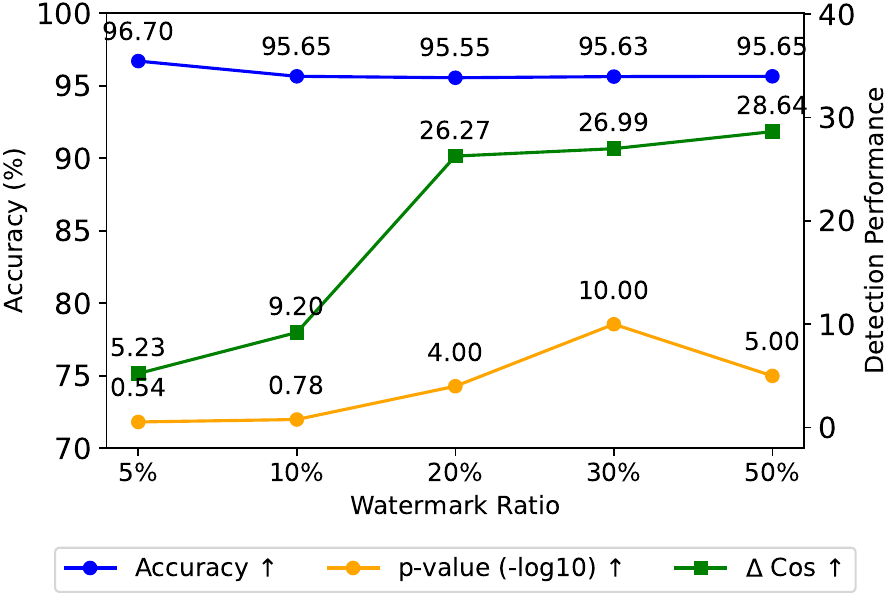}
        \caption{\ourattack}
    \end{subfigure}
    \hfill
    \begin{subfigure}[b]{0.32\textwidth}
        \includegraphics[width=\textwidth]{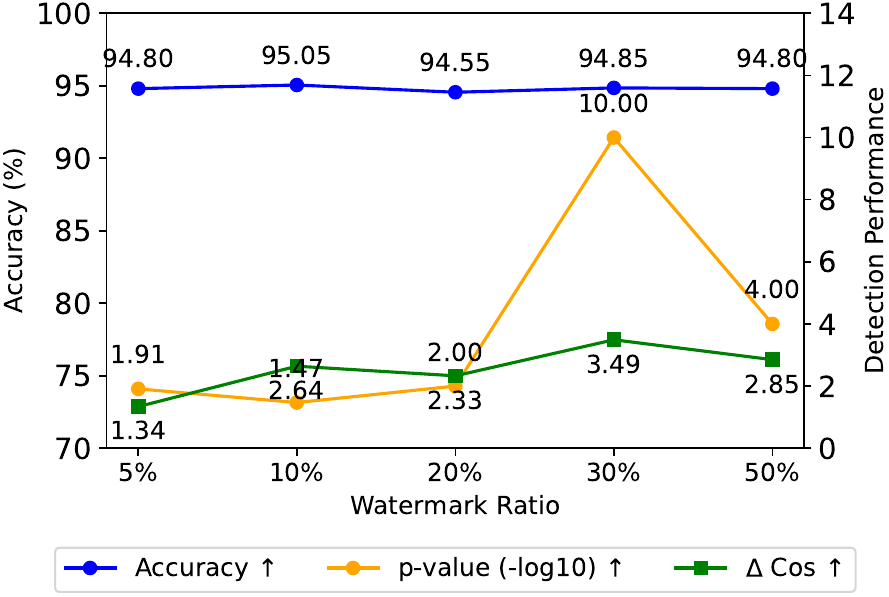}
        \caption{\dimshift}
    \end{subfigure}
    \caption{Impact of the proportion of watermarked regions $\alpha$ under different attacks on the Enron dataset.}
    \label{fig:two_images_enronb}
\end{figure*}

\subsection{Ablation Study}
\paragraph{The Necessity of Dimensionality Reduction}
We explore the necessity of introducing dimensionality reduction.
As shown in Table \ref{table:necessity-pca}, using PCA generally results in better performance compared to not using PCA.
According to the results in Figure~\ref{fig:region_distribution}, we speculate that this is due to the uneven distribution of data in the SST2 dataset, and dimensionality reduction helps make the data distribution more uniform.

Under CSE attacks, the performance of RegionMarker without PCA significantly declines, while the performance of RegionMarker with PCA only slightly decreases.
This highlights the necessity of introducing dimensionality reduction, and that dividing the space after dimensionality reduction and embedding watermark vectors makes it more covert and harder to break.

\paragraph{The Necessity of Multiple Watermark Embeddings}
We also investigate the necessity of using multiple watermark embeddings. 
When assigning the same watermark embedding to all trigger regions, we observe a noticeable drop in detection performance, as shown in Table~\ref{table:necessity-pca}. 
In particular, under the CSE attack, the watermark becomes ineffective on the SST-2 dataset.
This is because a single watermark can be easily identified and removed, whereas multiple embeddings increase the difficulty of removal, highlighting the necessity of using multiple watermark embeddings.

\begin{figure*}[!htbp]
    \centering
    \begin{subfigure}[b]{0.32\textwidth}
        \includegraphics[width=\textwidth]{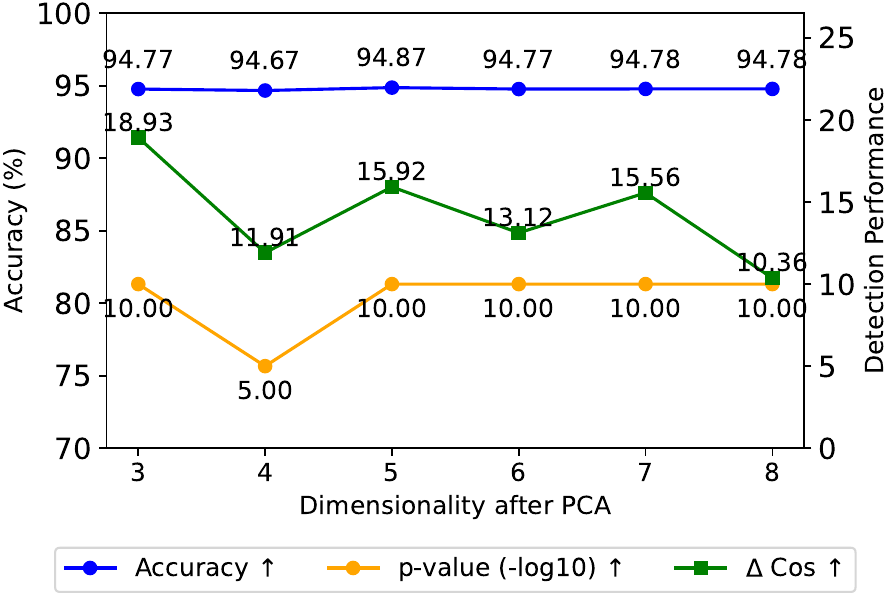}
        \caption{No Attack}
    \end{subfigure}
    \hfill
    \begin{subfigure}[b]{0.32\textwidth}
        \includegraphics[width=\textwidth]{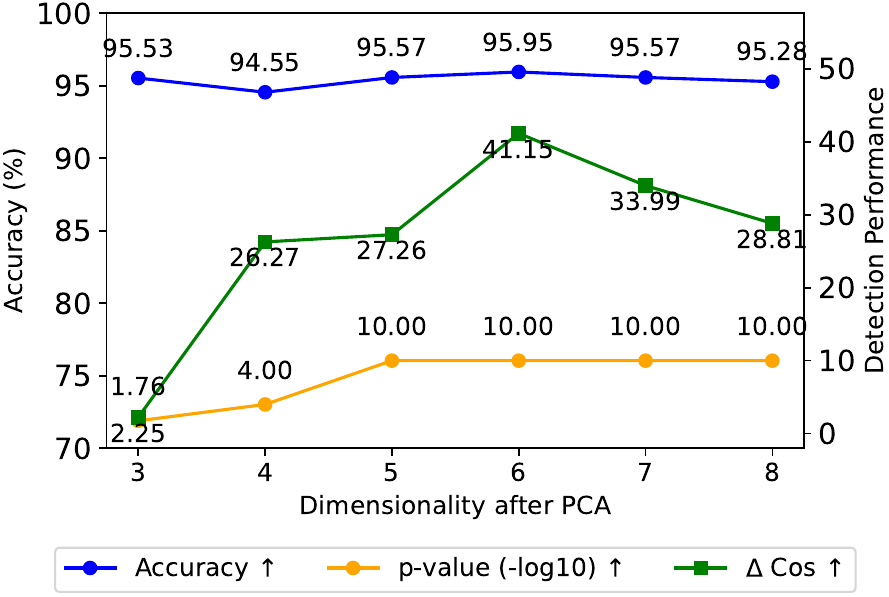}
        \caption{\ourattack}
    \end{subfigure}
    \hfill
    \begin{subfigure}[b]{0.32\textwidth}
        \includegraphics[width=\textwidth]{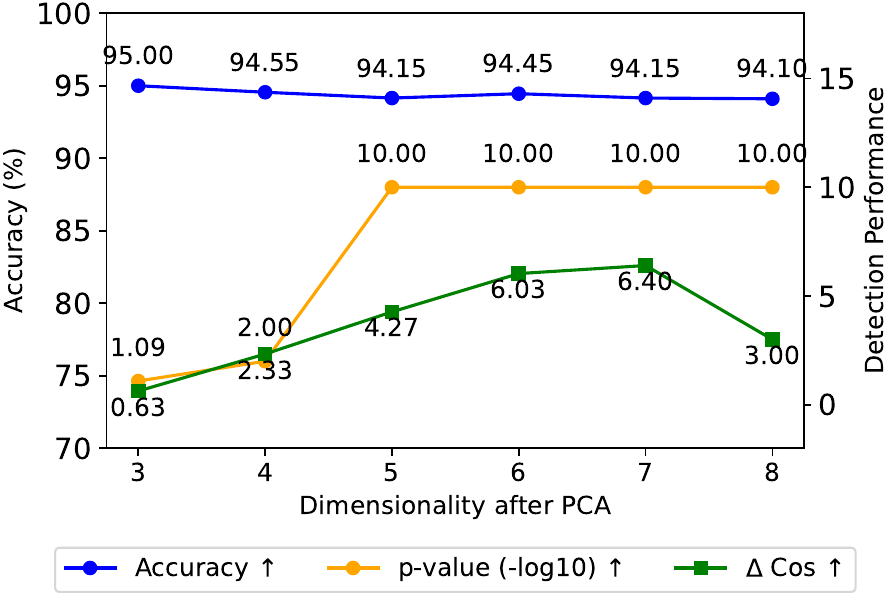}
        \caption{\dimshift}
    \end{subfigure}
    \caption{Impact of dimensionality after PCA under different attacks on the Enron dataset.}
    \label{fig:two_images_enronw}
\end{figure*}


\subsection{Hyper-parameter Analysis}
\label{sec:hyper-parameter analysis}
We explore the effects of watermark region ratio $\alpha$ and dimensionality after PCA on \swd using the Enron dataset.
We select two of the most challenging attack strategies, \textit{i.e.}, CSE and dimension-shift attacks, for evaluation.

Figure \ref{fig:two_images_enronb} shows that as the watermark region ratio $\alpha$ increases, the detection performance under different attacks also improves.
This is because we also use a conservative detection strategy following \citep{shetty-etal-2024-warden}, where the best watermark vector from all the watermark vectors is used to determine infringement.
However, we maintain a relatively low watermark ratio, namely 20\%.

Figure \ref{fig:two_images_enronw} shows that as the dimensionality after PCA increases, the $\Delta_{cos}$ exhibits an overall decreasing trend in the absence of attacks, while showing an overall increasing trend under different attacks.
This is because, with the increase in regions, the number of watermark embeddings gradually grows, and the sample size in each watermark region decreases, making it difficult for the extraction model to learn the watermark embeddings.
However, the increase in the number of watermark embeddings simultaneously raises the difficulty for attackers to successfully remove or bypass the watermarks.
Considering these factors, we select a watermark dimension of 4 and a watermark proportion of $20\%$.

For the watermark strength $\lambda$, we follow previous methods and set it to 0.2 to maintain a relatively low level.
Although a higher strength can enhance detection performance, it may compromise embedding quality.
To balance robustness and fidelity, we adopt this moderate setting.
See Appendix for results on other datasets.

\section{Conclusion}
We first reveal that current watermarking techniques for EaaS are unable to defend against existing attacks.
To this end, we propose a region-triggered semantic watermarking framework, which utilizes semantics rather than trigger words as triggers, and is capable of effectively defending against existing attack strategies.
Experiments show that our defense method provides comprehensive and effective defense against existing attacks. 

\section{Acknowledgments}
We would like to thank the anonymous reviewers for their insightful comments. 
This work is supported by the JiangSu Natural Science Foundation under Grant No. BK20251989; the National Natural Science Foundation of China under Grants Nos. 62172208, 62441225, 61972192; the Fundamental Research Funds for the Central Universities under Grant No. 14380001. 
This work is partially supported by Collaborative Innovation Center of Novel Software Technology and Industrialization.

\bibliography{custom}


\setlength{\leftmargini}{20pt}
\makeatletter
\def\@listi{\leftmargin\leftmargini \topsep .5em \parsep .5em \itemsep .5em}
\def\@listii{\leftmargin\leftmarginii \labelwidth\leftmarginii \advance\labelwidth-\labelsep \topsep .4em \parsep .4em \itemsep .4em}
\def\@listiii{\leftmargin\leftmarginiii \labelwidth\leftmarginiii \advance\labelwidth-\labelsep \topsep .4em \parsep .4em \itemsep .4em}
\makeatother

\setcounter{secnumdepth}{0}
\renewcommand\thesubsection{\arabic{subsection}}
\renewcommand\labelenumi{\thesubsection.\arabic{enumi}}

\newcounter{checksubsection}
\newcounter{checkitem}[checksubsection]

\newcommand{\checksubsection}[1]{%
  \refstepcounter{checksubsection}%
  \paragraph{\arabic{checksubsection}. #1}%
  \setcounter{checkitem}{0}%
}

\newcommand{\checkitem}{%
  \refstepcounter{checkitem}%
  \item[\arabic{checksubsection}.\arabic{checkitem}.]%
}
\newcommand{\question}[2]{\normalcolor\checkitem #1 #2 \color{blue}}
\newcommand{\ifyespoints}[1]{\makebox[0pt][l]{\hspace{-15pt}\normalcolor #1}}

\clearpage
\appendix
\setcounter{page}{1}

\section{Appendix}



\begin{table*}[t]  \small
\centering

    \setlength{\tabcolsep}{4pt}
    \begin{tabular}{clccccccc}
    \toprule
    \multirow{3}{*}{\textbf{Defend}} & \multirow{3}{*}{\textbf{Attack}} & \multicolumn{2}{c}{\textbf{Task Performance}} & \multicolumn{3}{c}{\textbf{Detection Performance}} & \multirow{3}{*}{\textbf{COPY?}} \\
    \cmidrule(lr){3-4} \cmidrule(lr){5-7}
    {} & {} & \textbf{ACC.(\%)} & \textbf{$F_1$-score}\xspace & {\textbf{$p$-value} $\downarrow$} & {$\Delta_{cos}(\%) \uparrow$} & {$\Delta_{l2}(\%) \downarrow$}\\
    \midrule
    \multirow{6}{*}{\prevWM} & No Attack & 77.30$\pm$0.05 & 51.65$\pm$0.03 & {$< 10^{-3}$} & 1.83$\pm$0.06 & -3.65$\pm$0.11 & \cmark\\
    & + \ourattack & 76.30$\pm$0.24 & 50.10$\pm$0.31 & {$< 10^{-5}$} & 5.57$\pm$0.31 & -11.14$\pm$0.63 & \cmark\\
    
     & + \para & 75.29$\pm$0.11 & 50.22$\pm$0.11 & $> 0.06$ & 0.79$\pm$0.28 & -1.57$\pm$0.55 & \xmark\\
     & + \gpt & 76.83$\pm$0.24 & 51.41$\pm$0.47 & $> 0.07$ & 1.01$\pm$0.31 & -2.02$\pm$0.62 & \xmark\\
     & + \dimshift & {77.26$\pm$0.05} & {51.54$\pm$0.14} & {$< 10^{-5}$} & 3.84$\pm$0.09 & -7.68$\pm$0.17 & \cmark\\
     & + \Dimdelete & {76.97$\pm$0.13} & {50.98$\pm$0.09} & {$< 10^{-5}$} & 3.77$\pm$0.07 & -7.54$\pm$0.14 & \cmark\\
    \midrule
    \multirow{6}{*}{\espeW} & No Attack & 77.30$\pm$0.10 & 50.29$\pm$0.11 & {$< 10^{-8}$} & 8.68$\pm$0.24 & -17.36$\pm$0.47 & \cmark\\
    
     & + \ourattack & 75.48$\pm$0.18 & 50.86$\pm$0.19 & {$< 10^{-11}$} & 72.14$\pm$2.16 & -144.28$\pm$4.31 & \cmark\\
    
     & + \para & 77.29$\pm$0.03 & 51.21$\pm$0.02 & ${< 0.01}$ & {1.13$\pm$0.01} & {-2.26$\pm$0.03} & \cmark\\
     & + \gpt & 77.34$\pm$0.18 & 51.68$\pm$0.03 & ${> 0.39}$ & {0.79$\pm$0.42} & {-1.58$\pm$0.85} & \xmark\\
     & + \dimshift & {77.22$\pm$0.02} & {51.31$\pm$0.18} & {$< 10^{-3}$} & 2.37$\pm$0.11 & -4.74$\pm$0.22 & \cmark\\
     & + \Dimdelete & {77.06$\pm$0.01} & {50.96$\pm$0.15} & {$< 10^{-3}$} & 2.57$\pm$0.18& -5.15$\pm$0.37 & \cmark\\
    \midrule
    \multirow{6}{*}{\wet} & No Attack & 76.88$\pm$0.05 & 50.85$\pm$0.22 & {$< 10^{-10}$} & 84.78$\pm$2.21 & -169.59$\pm$4.42 & \cmark\\
    & + \ourattack & 75.11$\pm$0.08 & 50.25$\pm$0.47 & {$< 10^{-10}$} & 22.47$\pm$2.71 & -44.94$\pm$5.43 & \cmark\\
     & + \para & {77.05$\pm$0.11} & {50.91$\pm$0.03} & {$< 10^{-10}$} & 87.98$\pm$0.21 &-175.96$\pm$0.42 & \cmark\\
      & + \gpt & {76.83$\pm$0.30} & {50.61$\pm$0.42} & {$< 10^{-10}$} & 87.43$\pm$0.77 &-174.87$\pm$1.55 & \cmark\\
     & + \dimshift & 76.80$\pm$0.12 & 50.76$\pm$0.16 & $> 0.27$ & -2.53$\pm$2.41 &  5.07$\pm$4.81 & \xmark\\
     & + \Dimdelete &- & - & - & - &  - & \xmark\\
    \midrule
    \multirow{6}{*}{\shortstack{\textbf{RegionMarker} \\ \textbf{(Ours)}}} & No Attack & 77.19$\pm$0.10 & 51.49$\pm$0.18 & $< 10^{-10}$ & 15.67$\pm$2.99 & -31.34$\pm$5.97 & \cmark\\
    & + \ourattack & 75.39$\pm$0.16 & 50.28$\pm$0.31 & {$<10^{-3}$} & 14.15$\pm$7.05 & -28.30$\pm$14.09 & \cmark\\
     & + \para & {76.50$\pm$0.08} & {50.48$\pm$0.13} & {$<10^{-3}$} & 9.34$\pm$0.23 &-18.69$\pm$0.47 & \cmark\\
     & + \gpt & {76.69$\pm$0.25} & {52.02$\pm$0.14} & {$<10^{-3}$} & 8.02$\pm$1.44 &-16.04$\pm$2.89 & \cmark\\
     & + \dimshift & 77.20$\pm$0.05 & 51.47$\pm$0.25 & {$<10^{-4}$} & 4.00$\pm$0.71 &  -8.00$\pm$1.42 & \cmark\\
     & + \Dimdelete & 76.88$\pm$0.12 & 50.93$\pm$0.27 & {$<10^{-3}$} & 2.67$\pm$0.20 &  -5.33$\pm$0.40 & \cmark\\
    \bottomrule
    \end{tabular}
    \caption{Performance of different methods on the MIND dataset.} 
    \label{table:mind-attack-performance}
\end{table*}

\begin{table*}[t]  \small
\centering

    \setlength{\tabcolsep}{4pt}
    \begin{tabular}{clcccccc}
    \toprule
    \multirow{3}{*}{\textbf{Defend}} & \multirow{3}{*}{\textbf{Attack}} & \multicolumn{2}{c}{\textbf{Task Performance}} & \multicolumn{3}{c}{\textbf{Detection Performance}} & \multirow{3}{*}{\textbf{COPY?}} \\
    \cmidrule(lr){3-4} \cmidrule(lr){5-7}
    {} & {} & \textbf{ACC.(\%)} & \textbf{$F_1$-score}\xspace & \textbf{{$p$-value $\downarrow$}} & {$\Delta_{cos}(\%) \uparrow$} & {$\Delta_{l2}(\%) \downarrow$}\\
    \midrule
    \multirow{6}{*}{\prevWM} & No Attack & 93.62$\pm$0.09 & 93.61$\pm$0.09 & {$< 10^{-9}$} & 25.41$\pm$0.53 & -50.82$\pm$1.05 & \cmark\\
    & + \ourattack & 92.68$\pm$0.20 & 92.68$\pm$0.20 & ${> 0.27}$ & {0.33$\pm$0.32} & {-0.67$\pm$0.63} & \xmark\\
    
     & + \para & 92.32$\pm$0.04 & 93.30$\pm$0.04 & {$< 10^{-10}$} & 10.18$\pm$0.47 & -20.36$\pm$0.94 & \cmark\\
     & + \gpt & 92.68$\pm$0.14 & 92.68$\pm$0.14 & {$< 10^{-10}$} & 7.97$\pm$0.51 & -15.95$\pm$1.01 & \cmark\\
     & + \dimshift & {93.59$\pm$0.11} & {93.59$\pm$0.11} & {$< 10^{-10}$} & 7.94$\pm$0.19 & -15.95$\pm$0.39 & \cmark\\
     & + \Dimdelete & {93.41$\pm$0.13} & {93.41$\pm$0.13} & {$< 10^{-5}$} & 7.46$\pm$0.51 & -14.92$\pm$1.02 & \cmark\\
    \midrule
    \multirow{6}{*}{\espeW} & No Attack & 93.42$\pm$0.16 & 93.42$\pm$0.16 & {$< 10^{-11}$} & 9.59$\pm$0.74 & -19.19$\pm$1.49 & \cmark\\
    
     & + \ourattack & 93.00$\pm$0.12 & 93.00$\pm$0.12 & {$< 10^{-10}$} & 21.83$\pm$5.11 & -43.65$\pm$10.22 & \cmark\\
    
     & + \para & 93.57$\pm$0.06 & 93.56$\pm$0.06 & {$< 10^{-3}$} & 4.46$\pm$0.08 & {-8.91$\pm$0.15} & \cmark\\
     & + \gpt & 93.37$\pm$0.05 & 93.37$\pm$0.05 & {$< 10^{-3}$} & 1.73$\pm$0.08 & {-3.47$\pm$0.16} & \cmark\\
     & + \dimshift & 93.36$\pm$0.11 & 93.35$\pm$0.11 & {$< 10^{-3}$} & 7.16$\pm$0.28 & -14.31$\pm$0.56 & \cmark\\
     & + \Dimdelete & 93.10$\pm$0.11 & 93.10$\pm$0.11 & {$< 10^{-3}$} & 6.04$\pm$0.35 & -12.08$\pm$0.70 & \cmark\\
    \midrule
    \multirow{6}{*}{\wet} & No Attack & 92.96$\pm$0.14 & 92.96$\pm$0.14 & {$< 10^{-10}$} & 89.81$\pm$0.91 & -179.62$\pm$1.83 & \cmark\\
    & + \ourattack & 93.00$\pm$0.15 & 92.99$\pm$0.15 & {$< 10^{-10}$} & 28.64$\pm$1.19 & -57.27$\pm$2.38 & \cmark\\
     & + \para & {92.92$\pm$0.01} & {92.91$\pm$0.01} & {$< 10^{-10}$} & 90.07$\pm$0.99 &-180.15$\pm$1.98 & \cmark\\
     & + \gpt & {93.01$\pm$0.12} & {93.01$\pm$0.12} & {$< 10^{-10}$} & 89.39$\pm$1.05 & -178.79$\pm$2.10 & \cmark\\
     & + \dimshift & 93.11$\pm$0.12 & 93.11$\pm$0.12 & $> 0.63$ & -0.23$\pm$0.32  &  0.46$\pm$0.65 & \xmark\\
     & + \Dimdelete & - & - & - & -  &  - & \xmark\\
    \midrule
    \multirow{6}{*}{\shortstack{\textbf{RegionMarker} \\ \textbf{(Ours)}}} & No Attack & 93.64$\pm$0.09 & 93.63$\pm$0.09 & $< 10^{-10}$ & 19.71$\pm$2.36 & -39.42$\pm$4.72 & \cmark\\
    & + \ourattack & 93.21$\pm$0.09 & 93.21$\pm$0.09 & {$<10^{-4} $} & 21.62$\pm$11.40 & -43.26$\pm$22.80 & \cmark\\
     & + \para & 92.55$\pm$0.12 & 92.54$\pm$0.12 & $< 10^{-10}$
 & 16.32$\pm$3.73 & -32.65$\pm$7.46 & \cmark\\
 & + \gpt & 93.14$\pm$0.08 & 93.13$\pm$0.08 & $< 10^{-7}$
 & 7.52$\pm$3.17 & -15.04$\pm$6.35 & \cmark\\
     & + \dimshift & 93.58$\pm$0.07 & 93.57$\pm$0.07 & $< 10^{-5}$ & 4.76$\pm$1.34 &  -9.52$\pm$2.68 & \cmark\\
     & + \Dimdelete & 93.40$\pm$0.01 & 93.40$\pm$0.01 & $< 10^{-5}$ & 4.09$\pm$1.24 &  -8.18$\pm$2.48 & \cmark\\
    \bottomrule
    \end{tabular}
    \caption{Performance of different methods on the AG News dataset.} 
    \label{table:ag-attack-performance}
\end{table*}

\begin{table*}[t]  \small
\centering

    \begin{tabular}{clcccccc}
    \toprule
    \multirow{3}{*}{Dataset} & \multirow{3}{*}{Defend} & \multicolumn{2}{c}{Task Performance} & \multicolumn{3}{c}{Detection Performance} & \multirow{3}{*}{COPY?} \\
    \cmidrule(lr){3-4} \cmidrule(lr){5-7}
    {} & {} & ACC.(\%) & $F_1$-score\xspace & {$p$-value $\downarrow$} & {$\Delta_{cos}(\%) \uparrow$} & {$\Delta_{l2}(\%) \downarrow$}\\
    \midrule
    \multirow{4}{*}{SST2} & \prevWM & 92.85$\pm$0.27 & 92.85$\pm$0.27 & {$> 0.06$} & 0.60$\pm$0.36 & -1.21$\pm$0.72 & \xmark\\
    & \espeW & 92.23$\pm$0.32 & 92.23$\pm$0.32 & $> 0.08$ & 0.91$\pm$0.08 & -1.82$\pm$0.15 & \xmark\\
    
     & \wet & {93.31$\pm$0.30} & {93.31$\pm$0.30} & $< 10^{-10}$ & 89.62$\pm$1.12 &-179.25$\pm$2.24 & \cmark\\
     & \textbf{RegionMarker (Ours)} & 93.50$\pm$0.05 & 93.50$\pm$0.05 & $< 10^{-5}$ & 9.56$\pm2.44$ & -19.91$\pm4.89$ & \cmark\\
    \midrule
    \multirow{4}{*}{AG News} & \prevWM & 92.53$\pm$0.08 & 92.53$\pm$0.08 & {$< 10^{-10}$} & 11.21$\pm$1.01 & -22.41$\pm$2.02 & \cmark\\
    
     & \espeW & 93.44$\pm$0.15 & 93.44$\pm$0.15 & {$< 10^{-3}$} & 5.96$\pm$0.14 & {-11.91$\pm$0.29} & \cmark\\
    
     & \wet & {92.41$\pm$0.80} & {92.41$\pm$0.80} & {$< 10^{-10}$} & 90.03$\pm$0.78 & -180.06$\pm$1.55 & \cmark\\
     & \textbf{RegionMarker (Ours)} & 93.00$\pm$0.02 & 93.00$\pm$0.02 & $< 10^{-7}$
 & 14.51$\pm$0.65 & -29.03$\pm$1.30 & \cmark\\
     
    \midrule
    \multirow{4}{*}{Enron} & \prevWM & 94.07$\pm$0.08 & 94.07$\pm$0.08 & $< 0.01$ & 1.27$\pm$0.30 & -2.55$\pm$0.60 & \cmark\\
    & \espeW & 94.85$\pm$0.16 &  94.85$\pm$0.16 & ${> 0.41}$ & {0.49$\pm$0.16} & {-0.98$\pm$0.32} & \xmark\\
     & \wet & {94.32$\pm$0.10} & {94.32$\pm$0.10} & {$< 10^{-10}$} & 86.69$\pm$0.18 &-173.38$\pm$0.36 & \cmark\\
     & \textbf{RegionMarker (Ours)} & 93.93$\pm$0.17 & 93.93$\pm$0.17 & ${< 10^{-6}}$
 & ${4.80\pm1.09}$ & ${-9.60\pm2.17}$ & \cmark\\
    \midrule
    \multirow{4}{*}{MIND} & \prevWM & 77.00$\pm$0.09 & 51.74$\pm$0.29 & $> 0.06$ & 0.79$\pm$0.14 & -1.58$\pm$0.27 & \xmark\\
    & \espeW & 77.09$\pm$0.18 & 51.72$\pm$0.24 & ${< 0.01}$ & {1.39$\pm$0.10} & {-2.78$\pm$0.20} & \cmark\\
     & \wet & {76.42$\pm$0.11} & {50.01$\pm$0.25} & {$< 10^{-10}$} & 86.34$\pm$0.11 &-172.68$\pm$0.22 & \cmark\\
 & \textbf{RegionMarker (Ours)} & {77.04$\pm$0.04} & {51.82$\pm$0.11} & {$<10^{-3}$} & 5.98$\pm$0.07 &-11.96$\pm$0.13 & \cmark\\
 
    \bottomrule
    \end{tabular}
    \caption{Performance of the DIPPER attack across four datasets and four defense methods.} 
    \label{table:dipper-attack-performance}
\end{table*}

\subsection{Experimental Settings}
\label{appendix:experimental-settings}

\subsubsection{Implementation Details of Defend Baselines}
\label{appendix:defend-baselines}
Since \textbf{WARDEN} is an improved version of \textbf{EmbMarker}, we adopt \textbf{WARDEN} as one of the baselines in this work.
For \textbf{WARDEN} and \textbf{EspeW}, we set the trigger word frequency to the range of [0.5\%, 1\%], and the trigger set size for each watermark is 20. 
For \textbf{WARDEN}, we select two watermark embeddings.
For \textbf{EspeW}, we set the watermark ratio to 20\%.
For \textbf{WET}, we set the number of correlations to 25.

\subsubsection{Description of CSE Attack}
\label{appendix:cse-attack}
\textbf{CSE} first applies a clustering algorithm, such as K-Means \citep{ilprints778}, to group the embeddings within the dataset, and then introduces a benchmark model. 
Within each cluster, given a pair of samples, CSE performs pairwise evaluations by comparing the embeddings produced by the victim model and the benchmark model to identify suspicious samples. 
For the embeddings of the suspicious samples, CSE first applies singular value decomposition (SVD) \citep{golub1971singular} to extract the top $K$ principal components, and then iteratively removes these components using the Gram-Schmidt (GS) \citep{demmel1997applied} process.
We set the number of clusters to 20 and the number of principal components to 50.

\subsubsection{Description of Paraphrasing Attacks}
\label{appendix:paraphrasing-attack}
We adopt three paraphrasing methods: (1) an explicitly trained paraphraser \textbf{DIPPER}~\citep{krishna2023paraphrasing}; (2) a 1.3B variant of \textbf{NLLB}~\citep{nllbteam2022languageleftbehindscaling}, an open-source multilingual model that performs paraphrasing via round-trip translation; and (3) \textbf{gpt-4o-mini}, a compact yet capable language model that generates diverse paraphrases via zero-shot prompting.
For DIPPER, we set $lex = 40$ and $div = 40$ to achieve significant text modification while maintaining text quality. 
Due to space limitations, all results of the DIPPER attack are provided in Table \ref{table:dipper-attack-performance}.
For round-trip translation, we first translate English into French and then translate it back into English. 
For gpt-4o-mini, we set the maximum token length to 1000 and the temperature to 0.7. 
The following prompt is used for paraphrasing:

\begin{tcolorbox}
\textbf{PROMPT}: “You are a helpful assistant to rewrite the text. Rewrite the following text:”
\end{tcolorbox}

For each input, five different paraphrased texts are generated. 
These candidates are filtered based on their cosine similarity with the original input, discarding low-quality paraphrases. 
The embeddings of the remaining high-quality paraphrases are then averaged to obtain the final provider's embedding. 
The cosine similarity threshold is set to 80$\%$.

\subsubsection{Description of Dimension-perturbation Attacks}
\label{appendix:dimension-shift-attack}
We adopt two methods for dimension-perturbation attacks: one cyclically shifts the embedding dimensions by 100 positions, and the other truncates the embedding to retain only the first 1024 dimensions.
For example, let the original vector be $\textbf{v}=(v_1,v_2,...,v_d)$, where $v_i$ denotes the $i$-th dimension of $\textbf{v}$, and $d$ is the dimensionality of $\textbf{v}$. 
After applying circular shift, the resulting vector is $\textbf{v}'=(v_{d-99},v_{d-98},\cdots, v_{d-1}, v_{d}, v_{1}, v_{2}, \cdots,v_{d-100})$.
After applying truncation, the resulting vector is $\textbf{v}'=(v_1,v_2,...,v_{1024})$.
Both methods have minimal impact on the quality of the embeddings.

\subsection{Results on the MIND and AG News datasets}
\label{appendix:more-results}
We present the main results on the MIND and AG News datasets in Table \ref{table:mind-attack-performance} and Table \ref{table:ag-attack-performance}.

Experimental results also show that our method remains effective in defending against all attack strategies. 
Notably, paraphrasing attacks completely failed on the AG News dataset and partially failed on the MIND dataset. 
We speculate that this is because both datasets contain relatively long texts, making it difficult to fully remove the trigger words during paraphrasing. 
In the case of AG News, the longer text length may significantly hinder the removal of all triggers, and in some cases, paraphrasing may even introduce new trigger words, leading to complete failure of the attacks. 
For MIND, while the effect is not as strong, the text length still poses challenges to effective trigger word removal, allowing some methods to remain robust.

\begin{figure*}[t]
    \centering
    \begin{subfigure}[b]{0.32\textwidth}
        \includegraphics[width=\textwidth]{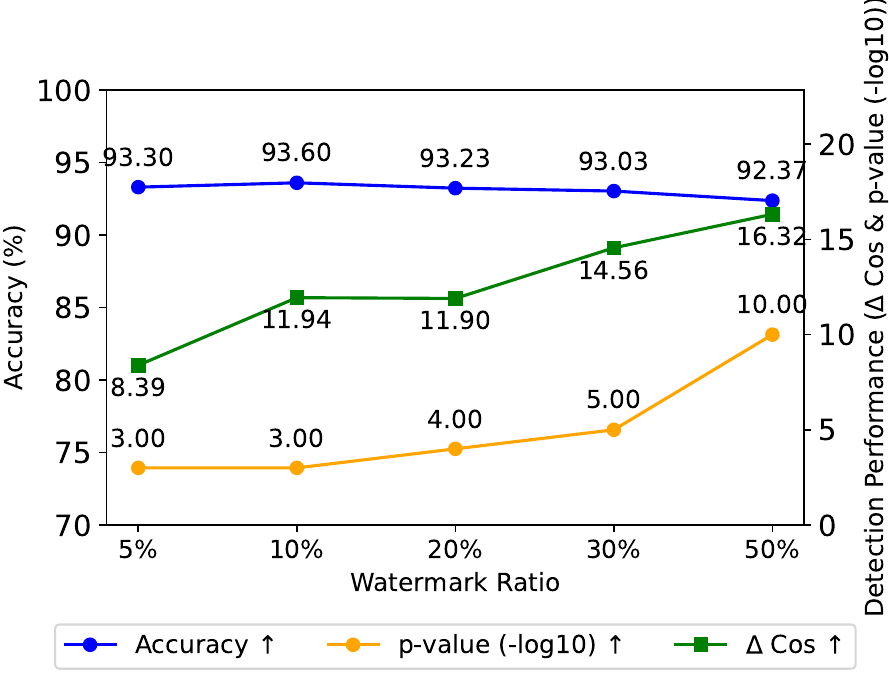}
        \caption{No Attack}
    \end{subfigure}
    \hfill
    \begin{subfigure}[b]{0.32\textwidth}
        \includegraphics[width=\textwidth]{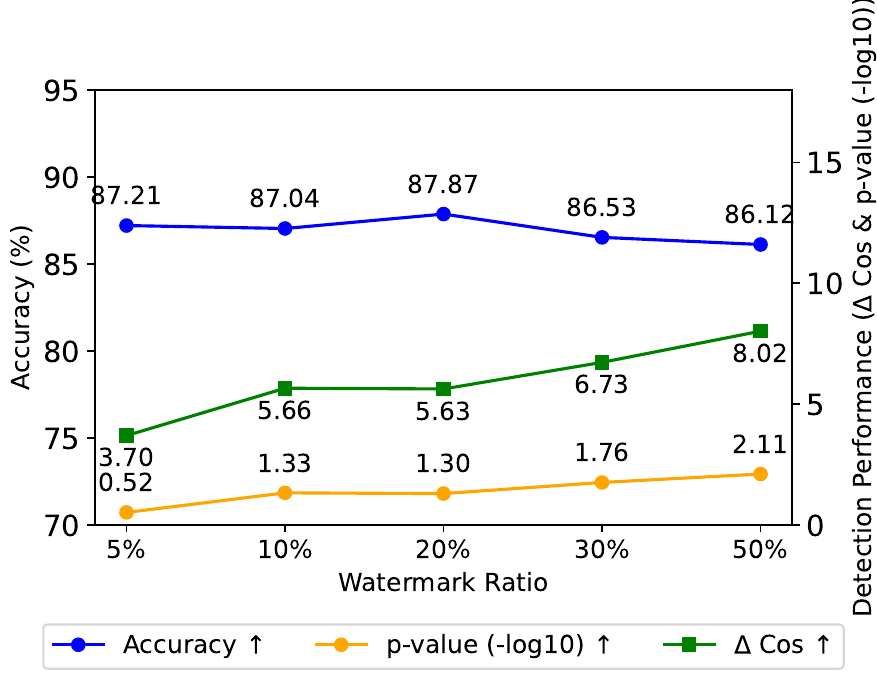}
        \caption{\ourattack}
    \end{subfigure}
    \hfill
    \begin{subfigure}[b]{0.32\textwidth}
        \includegraphics[width=\textwidth]{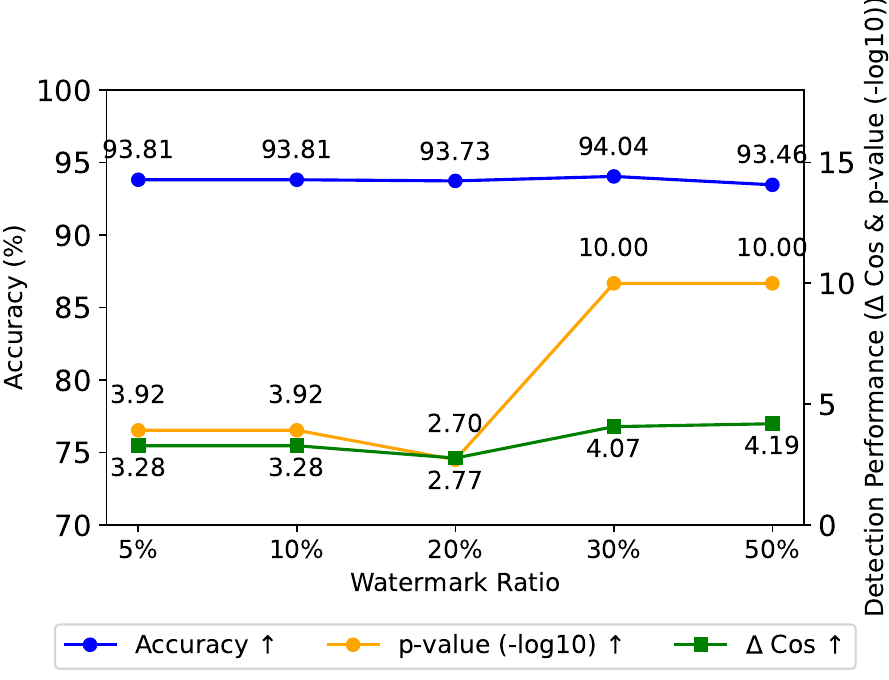}
        \caption{\dimshift}
    \end{subfigure}
    \caption{Impact of the proportion of watermarked regions $\alpha$ under different attacks on the SST2 dataset.}
    \label{fig:two_images_sst2b}
\end{figure*}

\begin{figure*}[h]
    \centering
    \begin{subfigure}[b]{0.32\textwidth}
        \includegraphics[width=\textwidth]{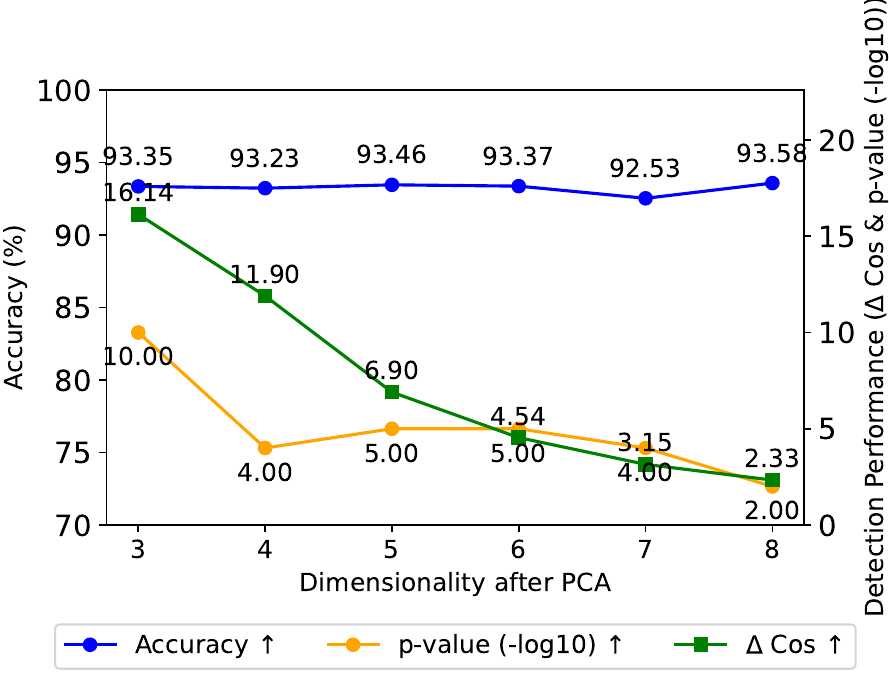}
        \caption{No Attack}
    \end{subfigure}
    \hfill
    \begin{subfigure}[b]{0.32\textwidth}
        \includegraphics[width=\textwidth]{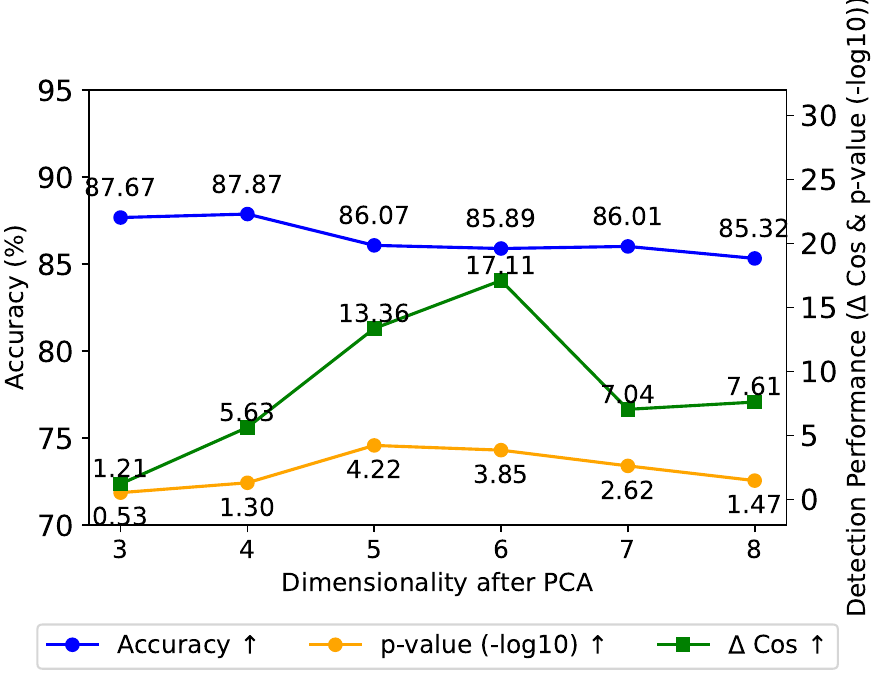}
        \caption{\ourattack}
    \end{subfigure}
    \hfill
    \begin{subfigure}[b]{0.32\textwidth}
        \includegraphics[width=\textwidth]{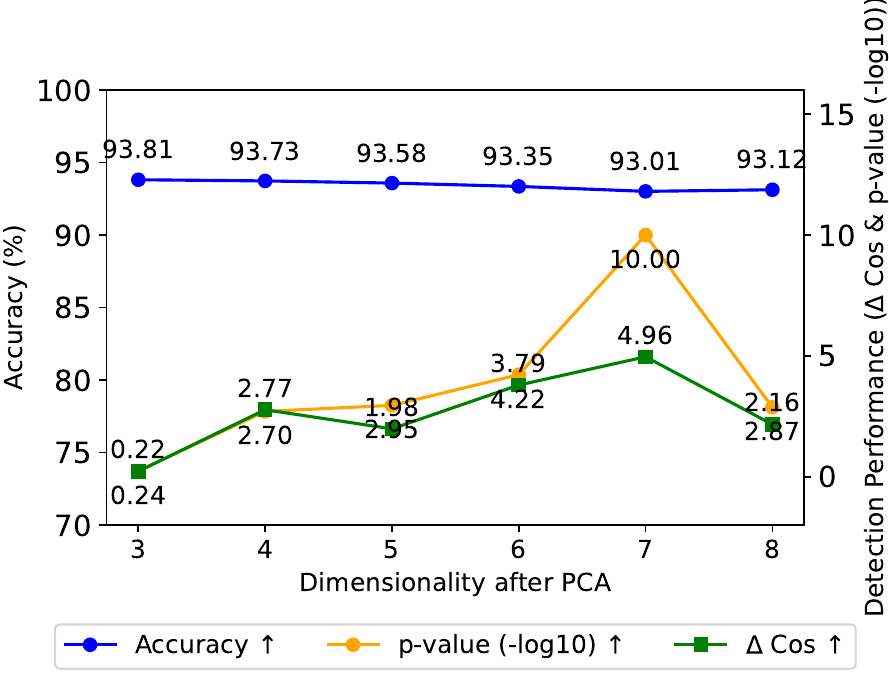}
        \caption{\dimshift}
    \end{subfigure}
    \caption{Impact of dimensionality after PCA under different attacks on the SST2 dataset.}
    \label{fig:two_images_sst2w}
\end{figure*}

\begin{figure*}[h]
    \centering
    \begin{subfigure}[b]{0.32\textwidth}
        \includegraphics[width=\textwidth]{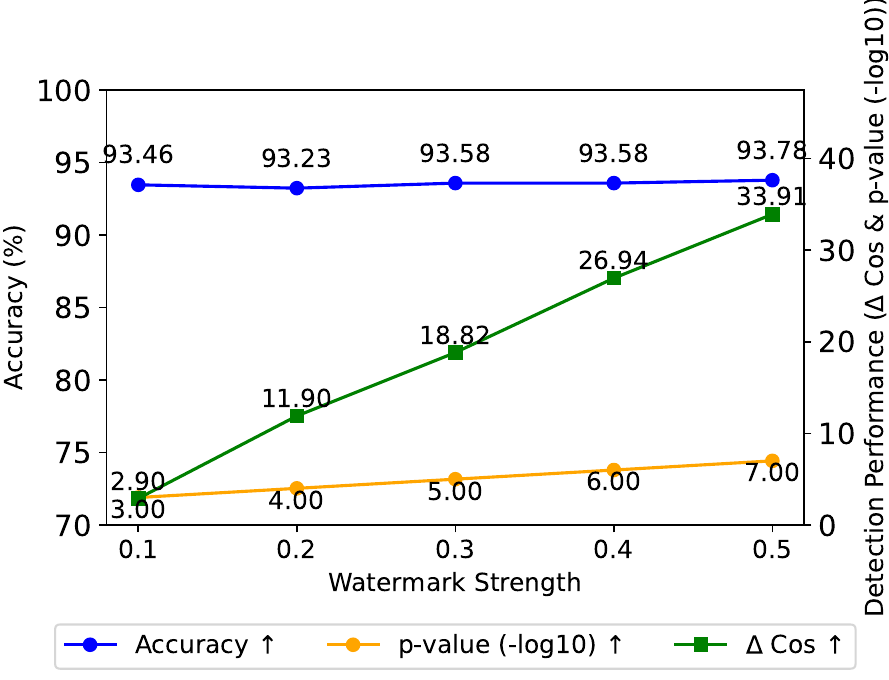}
        \caption{No Attack}
    \end{subfigure}
    \begin{subfigure}[b]{0.32\textwidth}
        \includegraphics[width=\textwidth]{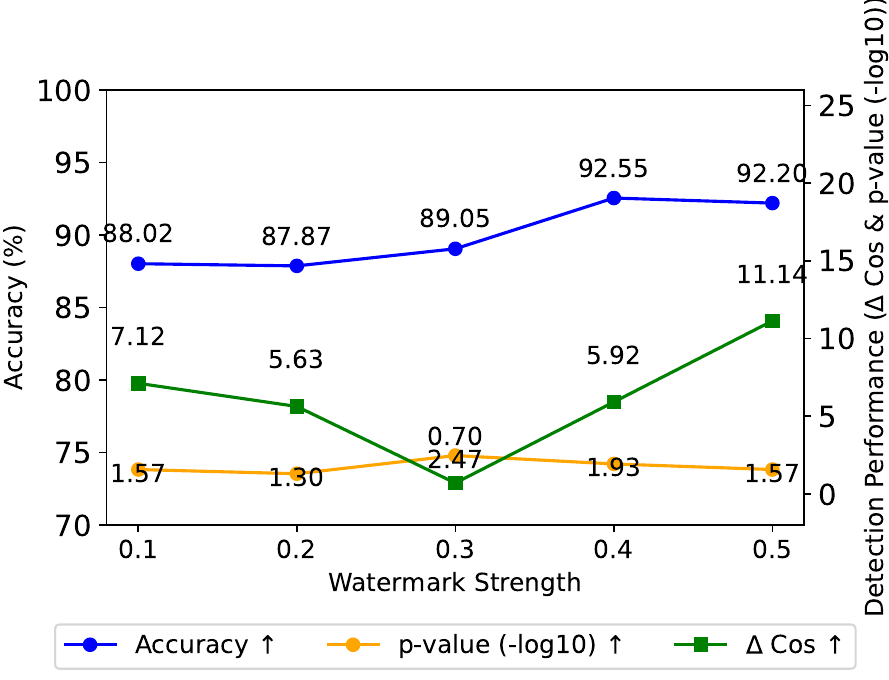}
        \caption{\ourattack}
    \end{subfigure}
    \begin{subfigure}[b]{0.32\textwidth}
        \includegraphics[width=\textwidth]{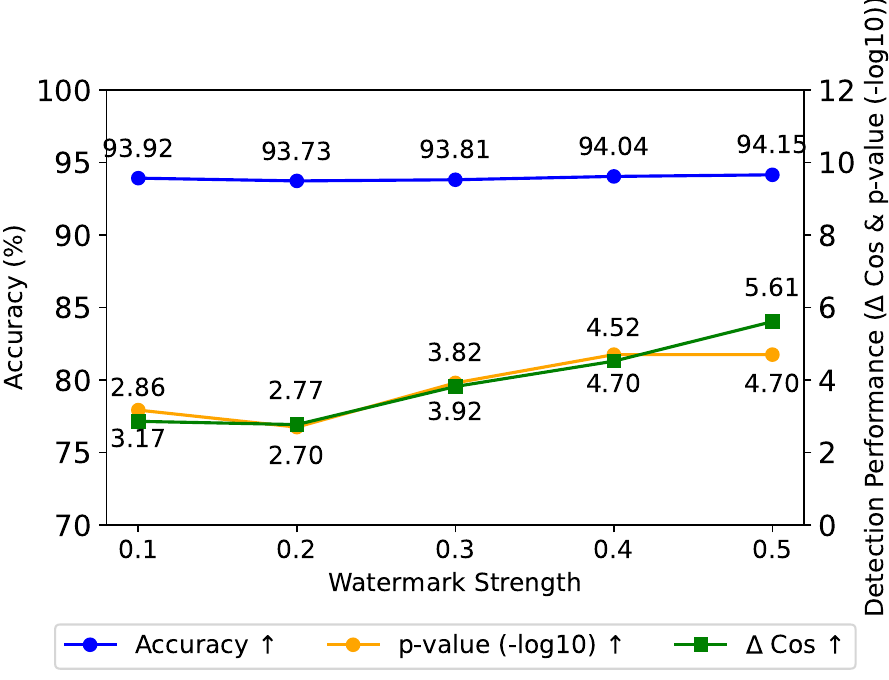}
        \caption{\dimshift}
    \end{subfigure}
    \caption{Impact of watermark strength $\lambda$ under different attacks on the SST2 dataset.}
    \label{fig:watermark_weight_sst2}
\end{figure*}

\begin{figure*}[t]
    \centering
    \begin{subfigure}[b]{0.47\textwidth}
        \includegraphics[width=\textwidth]{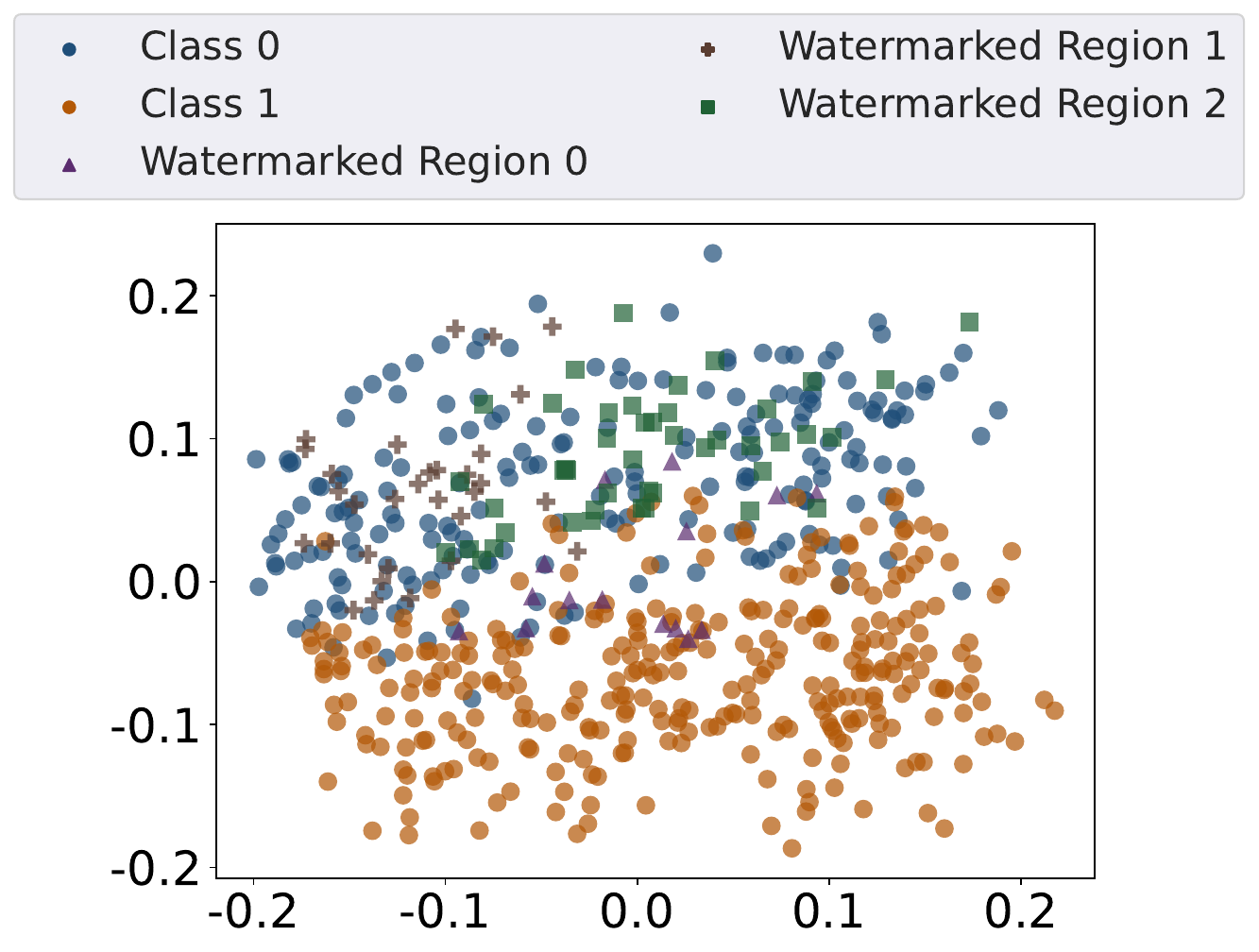}
        \caption{Original (Before Watermarking)}
    \end{subfigure}
    \begin{subfigure}[b]{0.47\textwidth}
        \includegraphics[width=\textwidth]{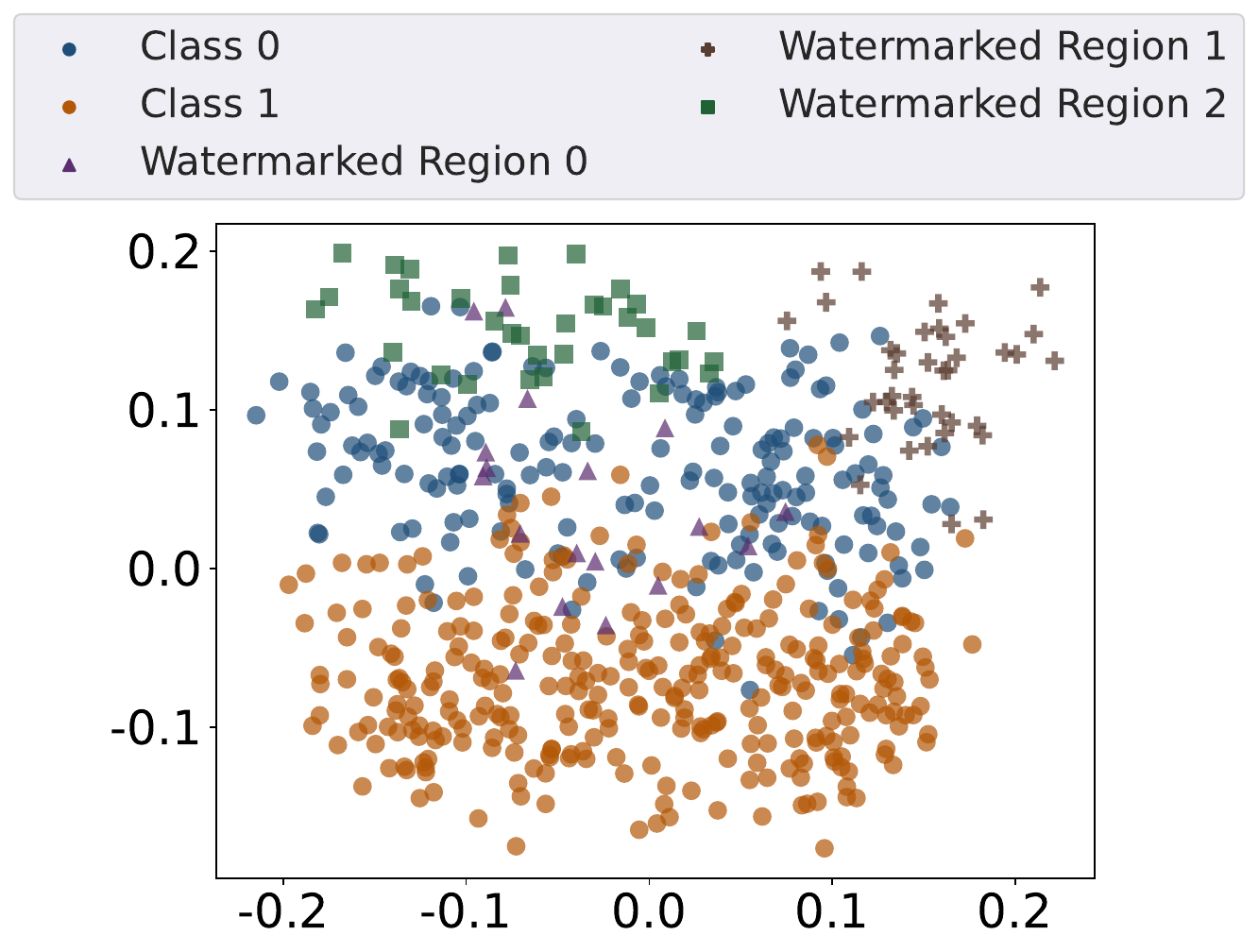}
        \caption{\swd (After Watermarking)}
    \end{subfigure}
    \caption{Visualization of the embeddings before and after watermarking on the SST-2 dataset. Embeddings in watermark regions will be added to the watermark embedding.}
    \label{fig:visualization}
\end{figure*}

WET also demonstrates strong robustness against paraphrasing attacks because it embeds watermarks using linear transformations, which is a text-independent watermarking approach. 
WET applies a linear transformation to all text embeddings, so regardless of how the input text is paraphrased, the resulting embeddings will still undergo the same transformation, making it difficult to evade detection.

\subsection{Results of DIPPER Attacks}
\label{appendix:dipper}
Due to space constraints, the results of the DIPPER attack on all datasets and defense methods are presented in Table \ref{table:dipper-attack-performance}. 
Consistent with earlier observations, WARDEN and EspeW fail to withstand the DIPPER attack, whereas RegionMarker and WET maintain strong robustness.

\subsection{Results under Dimension-wise Transformation Attacks}
To evaluate watermark robustness, we simulate three types of dimension-wise transformations: a random permutation of all dimensions, non-linear scaling via the tanh function, and non-orthogonal transformations implemented through random projection matrices.

As shown in Table \ref{table:dimension-wise-attack-performance}, our method remains robust under all three dimension-wise transformations and consistently enables reliable watermark detection.

\begin{table*}[t]  \small
\centering
    \setlength{\tabcolsep}{3pt}
    \begin{tabular}{clcccccc}
    \toprule
    \multirow{3}{*}{\textbf{Defend}} & \multirow{3}{*}{\textbf{Attack}} & \multicolumn{2}{c}{\textbf{Task Performance}} & \multicolumn{3}{c}{\textbf{Detection Performance}} & \multirow{3}{*}{\textbf{COPY?}} \\
    \cmidrule(lr){3-4} \cmidrule(lr){5-7}
    {} & {} & \textbf{ACC.(\%)} & \textbf{$F_1$-score\xspace} & \textbf{{$p$-value $\downarrow$}} & \textbf{{$\Delta_{cos}(\%) \uparrow$}} & \textbf{{$\Delta_{l2}(\%) \downarrow$}}\\
    \midrule
    \multirow{4}{*}{\shortstack{\textbf{RegionMarker} \\ \textbf{(Ours)}}} & No Attack & 93.23$\pm$0.36 & 93.23$\pm$0.36 & {$< 10^{-4}$} & 11.90$\pm$3.75 & -23.80$\pm$7.50 & \cmark\\
     & + Random Permutation of Dimensions & 93.46$\pm$0.24 & 93.46$\pm$0.24 & {$< 10^{-3}$} & 3.35$\pm$0.75 & -6.71$\pm$1.51 & \cmark\\
     & + Non-linear Scaling of Embedding Values & 93.64$\pm$0.26 & 93.64$\pm$0.26 & {$< 10^{-3}$} & 3.75$\pm$0.76 &  -7.50$\pm$1.52 & \cmark\\
     & + Non-orthogonal Transformations & 93.21$\pm$0.31 & 93.21$\pm$0.3 & {$< 10^{-3}$} & 3.22$\pm$0.82 &  -6.44$\pm$1.65 & \cmark\\
    \bottomrule
    \end{tabular}
    \caption{Performance of our proposed method under dimension-wise transformation attacks.}
    \label{table:dimension-wise-attack-performance}
\end{table*}

\subsection{Results of Fine-tuning Attacks}
To further assess the robustness of RegionMarker, we investigate the impact of fine-tuning attacks. 
In particular, models extracted from the SST-2 dataset are further fine-tuned on the clean Enron, AG News, and MIND datasets. 
As shown in Table \ref{table:modify-attack-performance}, the watermark remains clearly detectable after fine-tuning.

\begin{table*}[t]  \small
\centering
    \setlength{\tabcolsep}{3pt}
    \begin{tabular}{lcccc}
    \toprule
    \multirow{3}{*}{\textbf{Dataset}} & \multicolumn{3}{c}{\textbf{Detection Performance}} & \multirow{3}{*}{\textbf{COPY?}} \\
    \cmidrule(lr){2-4} 
    {}  & \textbf{{$p$-value $\downarrow$}} & \textbf{{$\Delta_{cos}(\%) \uparrow$}} & \textbf{{$\Delta_{l2}(\%) \downarrow$}}\\
    \midrule
     Enron & {$< 10^{-4}$} & 5.49$\pm$0.54 & -10.98$\pm$1.09 & \cmark\\
     AG News & {$< 10^{-4}$} & 3.79$\pm$0.75 & -7.59$\pm$1.51 & \cmark\\
     MIND & {$< 10^{-4}$} & 8.93$\pm$2.35 &  -17.86$\pm$4.69 & \cmark\\
    \bottomrule
    \end{tabular}
    \caption{Performance of our proposed method under fine-tuning attacks.}
    \label{table:modify-attack-performance}
\end{table*}

\subsection{Effectiveness Against Dimension-perturbation Attacks}
All watermarking strategies except WET (\textit{i.e.}, WARDEN, EspeW, and RegionMarker) can use the embedding of specific texts as the watermarks, and thus are resistant to dimension-perturbation attacks.
Specifically, the provider model can directly use the embedding $\textbf{e}_t$ of the target text $t$ as the watermark embedding to be injected into the original embedding $\textbf{e}_o$, resulting in the watermarked embedding $\textbf{e}_p$, as shown in Equation~\eqref{eq:3}. 
When an attacker obtains $\textbf{e}_p$, they may perturb it via dimension-perturbation attacks, causing both $\textbf{e}_p$ and the watermark embedding $\textbf{e}_t$ to become $\textbf{e}_p'$ and $\textbf{e}_t'$, respectively. 
These methods re-input the target text $t$ into the stealer's model to obtain an embedding $\textbf{e}_t''$, which can approximate $\textbf{e}_t'$, and then use it for watermark detection. 

In contrast, WET relies on a linear transformation matrix. 
If the attacker perturbs $\textbf{e}_p$ into $\textbf{e}_t'$, the inverse of the transformation matrix can no longer recover the original embedding, leading to failure of watermark detection.
In particular, under dimension-reduction attacks, the transformation matrix becomes completely unusable due to dimensional mismatch.

\subsection{Effects of Hyper-parameters}
\begin{figure*}[t]
    \centering
    \begin{subfigure}[b]{0.32\textwidth}
        \includegraphics[width=\textwidth]{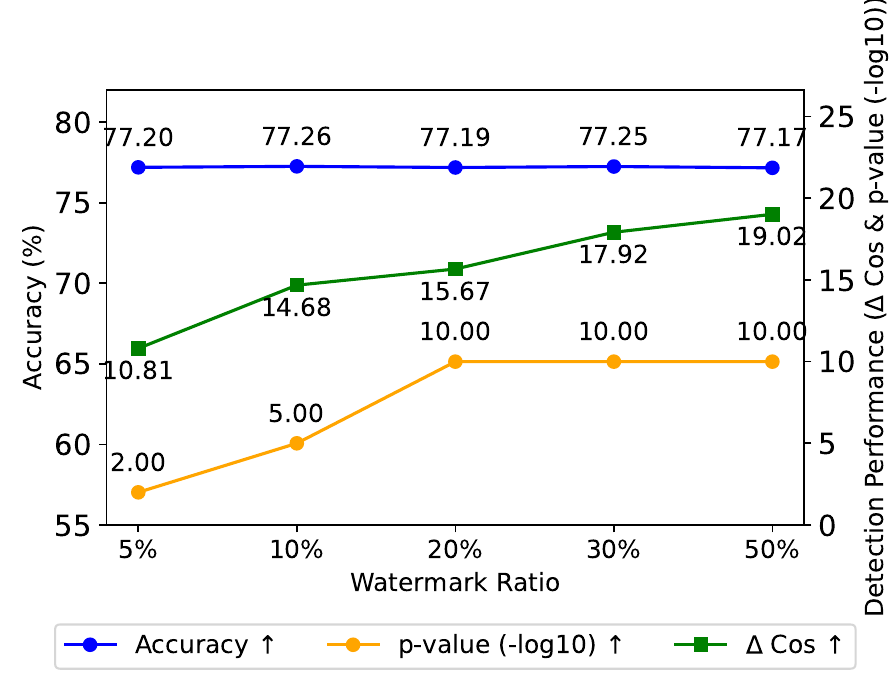}
        \caption{No Attack}
    \end{subfigure}
    \hfill
    \begin{subfigure}[b]{0.32\textwidth}
        \includegraphics[width=\textwidth]{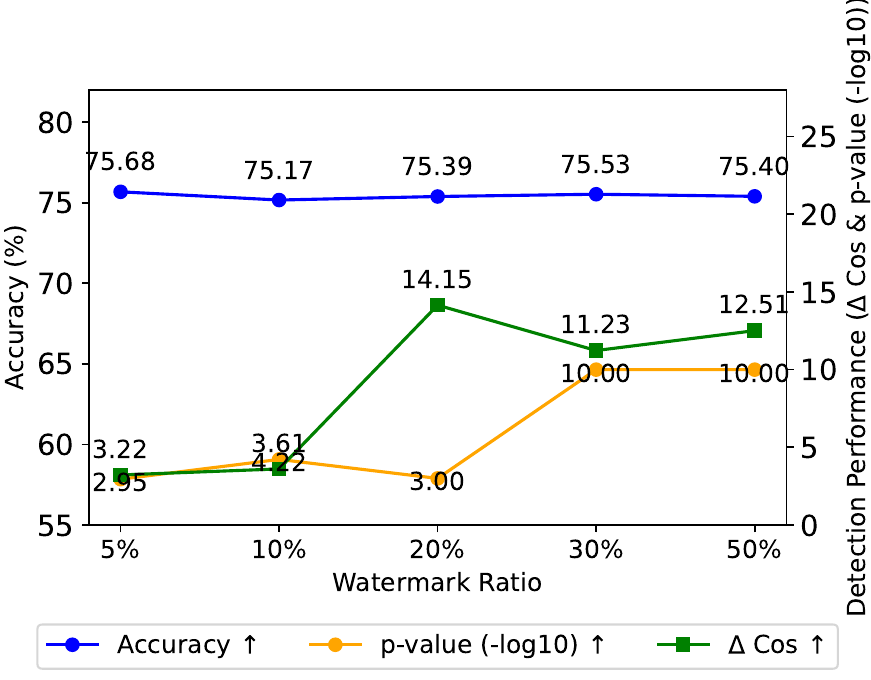}
        \caption{\ourattack}
    \end{subfigure}
    \hfill
    \begin{subfigure}[b]{0.32\textwidth}
        \includegraphics[width=\textwidth]{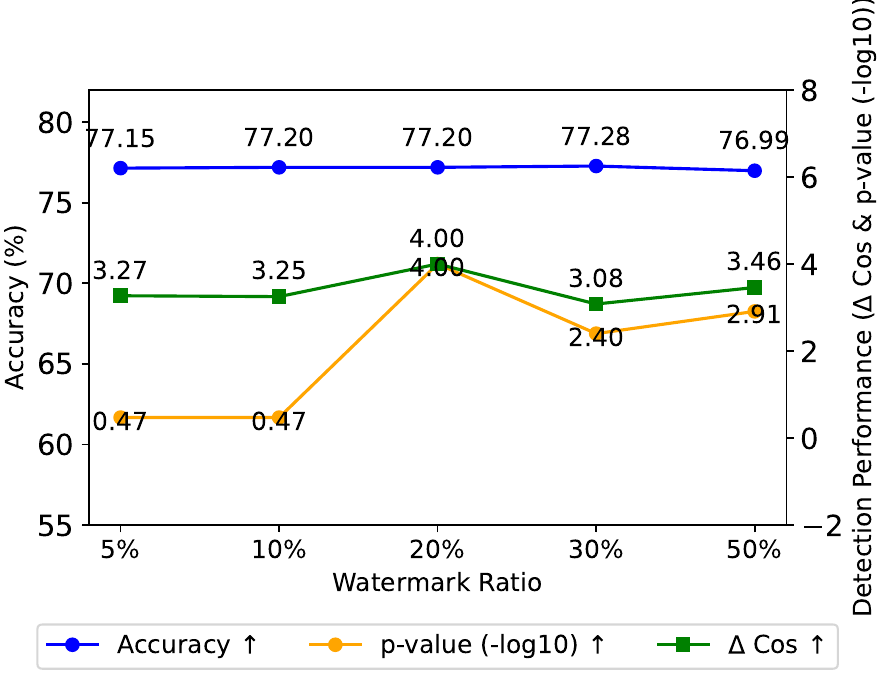}
        \caption{\dimshift}
    \end{subfigure}
    \caption{Impact of the proportion of watermarked regions $\alpha$ under different attacks on the MIND dataset.}
    \label{fig:two_images_mindb}
\end{figure*}

\begin{figure*}[h]
    \centering
    \begin{subfigure}[b]{0.32\textwidth}
        \includegraphics[width=\textwidth]{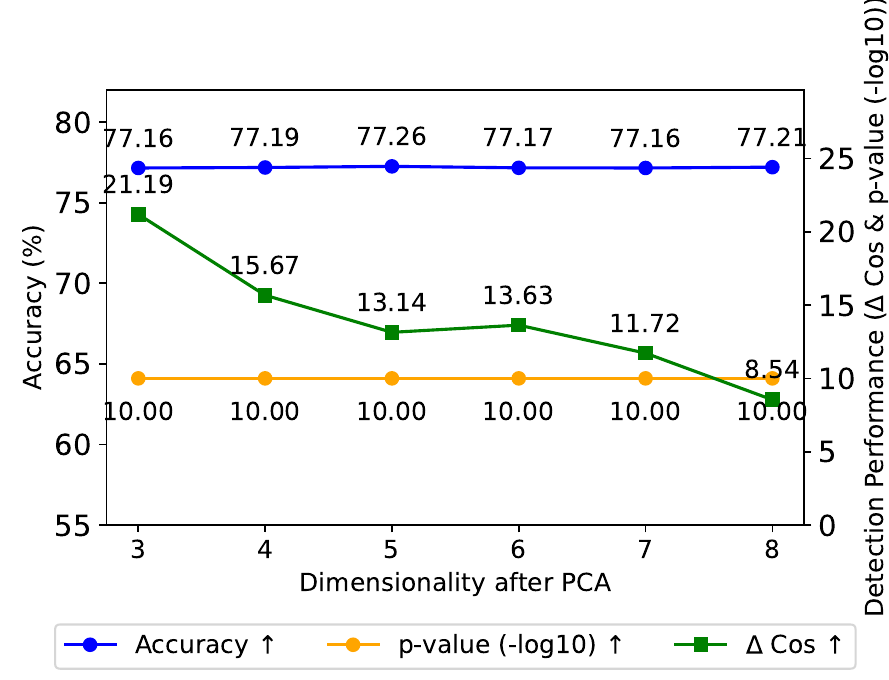}
        \caption{No Attack}
    \end{subfigure}
    \hfill
    \begin{subfigure}[b]{0.32\textwidth}
        \includegraphics[width=\textwidth]{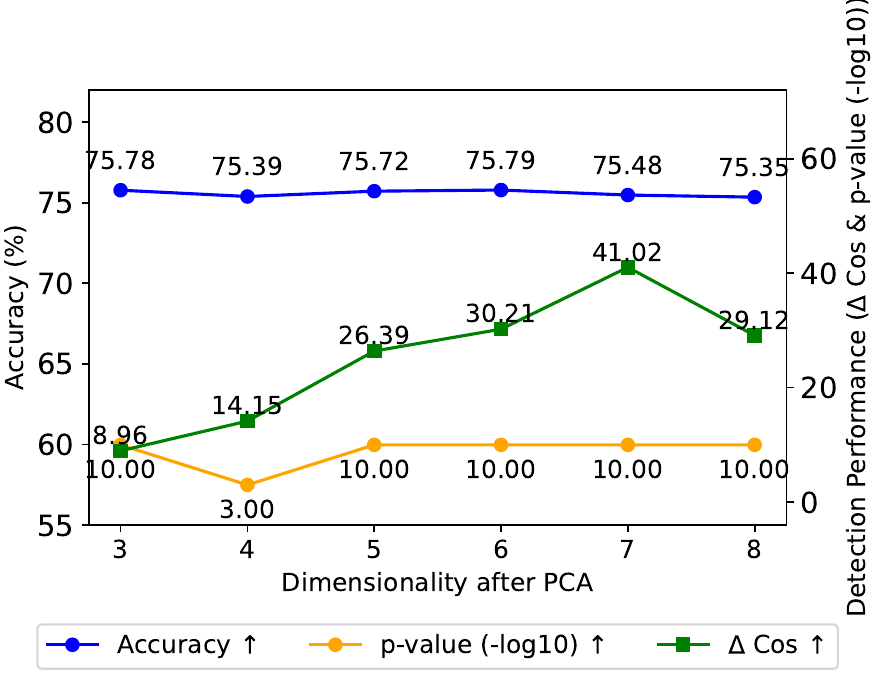}
        \caption{\ourattack}
    \end{subfigure}
    \hfill
    \begin{subfigure}[b]{0.32\textwidth}
        \includegraphics[width=\textwidth]{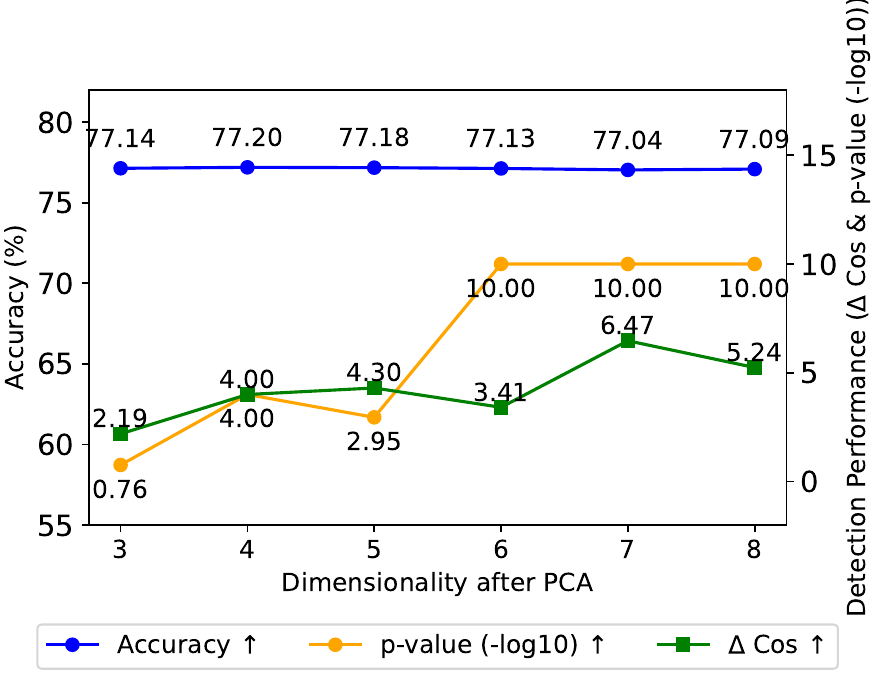}
        \caption{\dimshift}
    \end{subfigure}
    \caption{Impact of dimensionality after PCA under different attacks on the MIND dataset.}
    \label{fig:two_images_mindw}
\end{figure*}

\begin{figure*}[b]
    \centering
    \begin{subfigure}[b]{0.32\textwidth}
        \includegraphics[width=\textwidth]{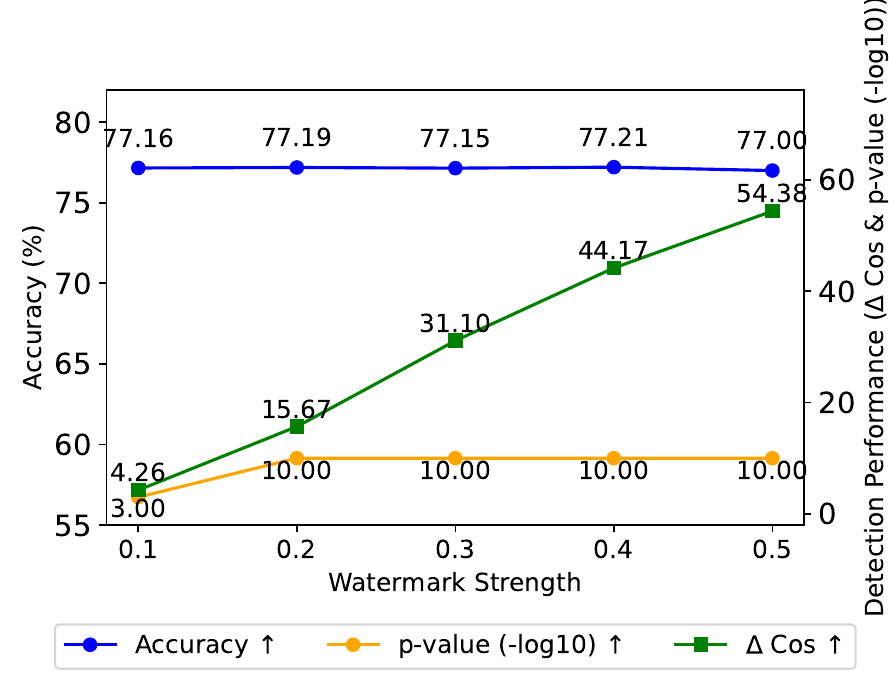}
        \caption{No Attack}
    \end{subfigure}
    \begin{subfigure}[b]{0.32\textwidth}
        \includegraphics[width=\textwidth]{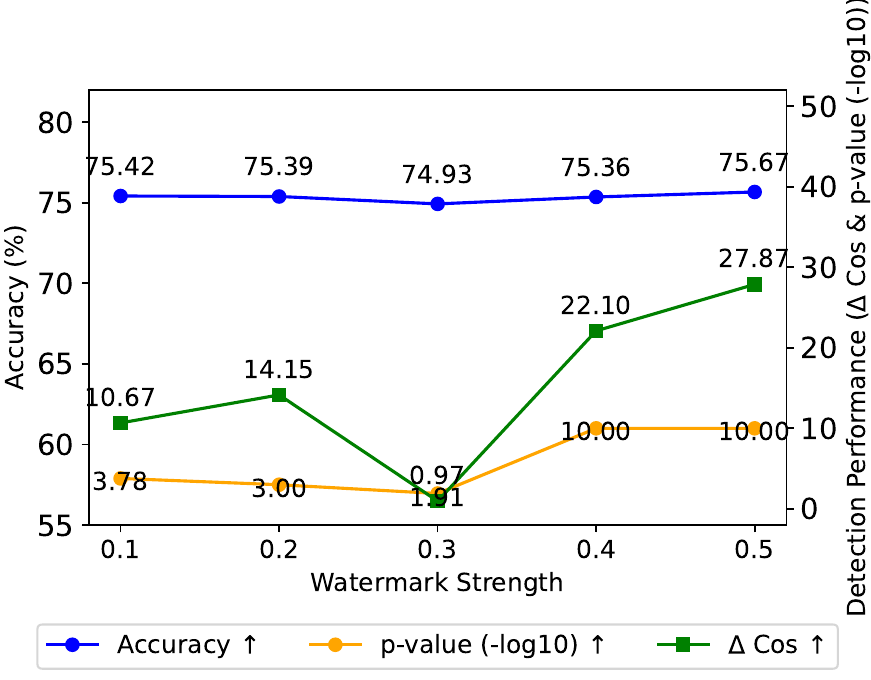}
        \caption{\ourattack}
    \end{subfigure}
    \begin{subfigure}[b]{0.32\textwidth}
        \includegraphics[width=\textwidth]{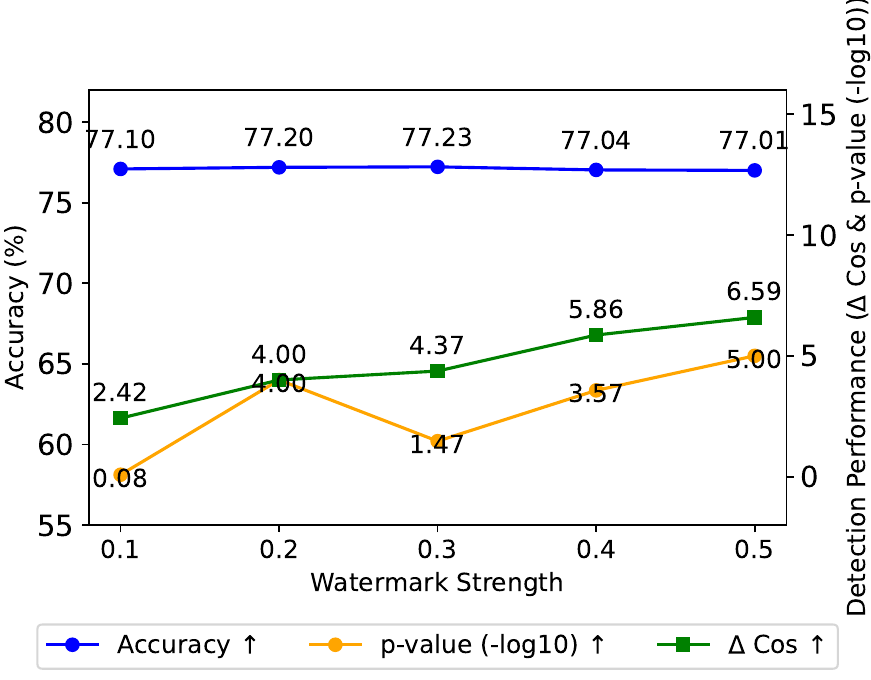}
        \caption{\dimshift}
    \end{subfigure}
    \caption{Impact of watermark strength $\lambda$ under different attacks on the MIND dataset.}
    \label{fig:watermark_weight_mind}
\end{figure*}

\begin{figure*}[t]
    \centering
    \begin{subfigure}[b]{0.32\textwidth}
        \includegraphics[width=\textwidth]{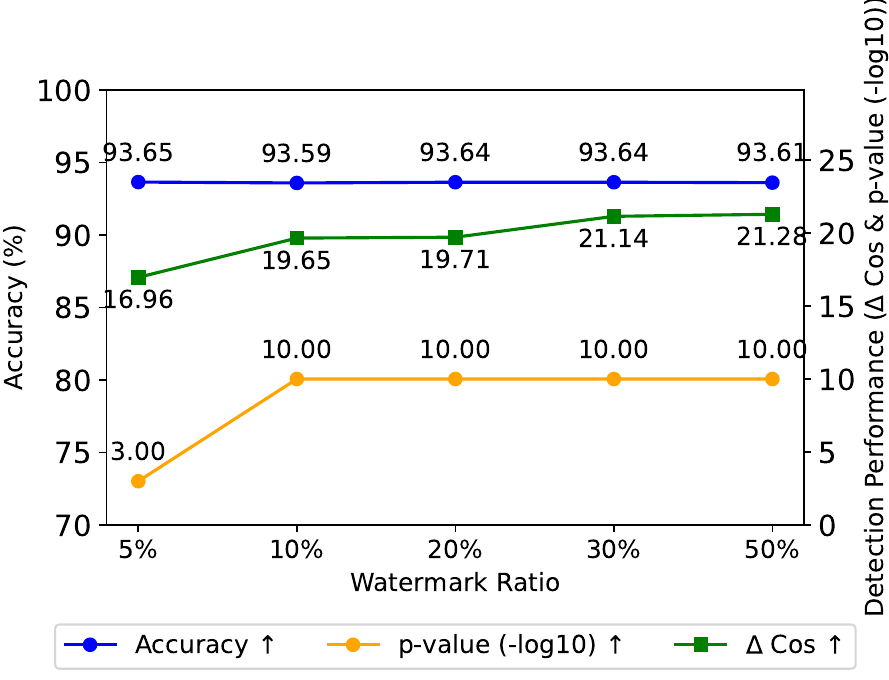}
        \caption{No Attack}
    \end{subfigure}
    \hfill
    \begin{subfigure}[b]{0.32\textwidth}
        \includegraphics[width=\textwidth]{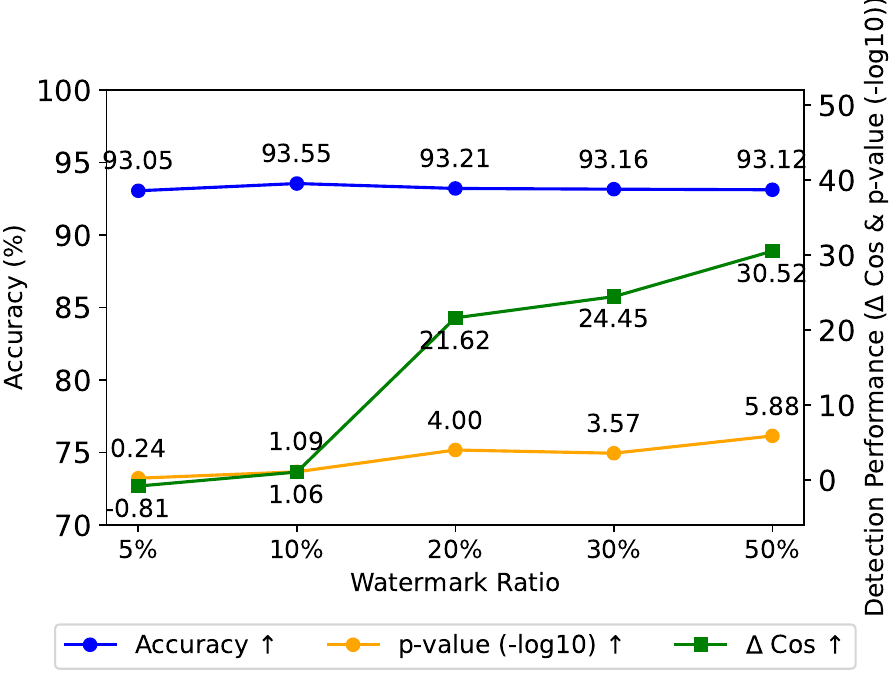}
        \caption{\ourattack}
    \end{subfigure}
    \hfill
    \begin{subfigure}[b]{0.32\textwidth}
        \includegraphics[width=\textwidth]{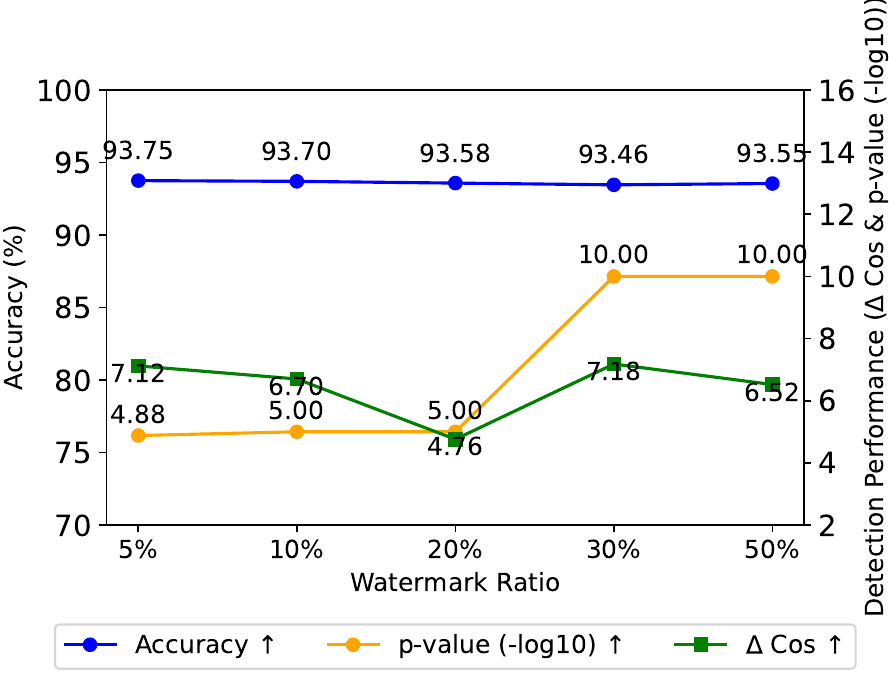}
        \caption{\dimshift}
    \end{subfigure}
    \caption{Impact of the proportion of watermarked regions $\alpha$ under different attacks on the AG News dataset.}
    \label{fig:two_images_agb}
\end{figure*}

\begin{figure*}[t]
    \centering
    \begin{subfigure}[b]{0.32\textwidth}
        \includegraphics[width=\textwidth]{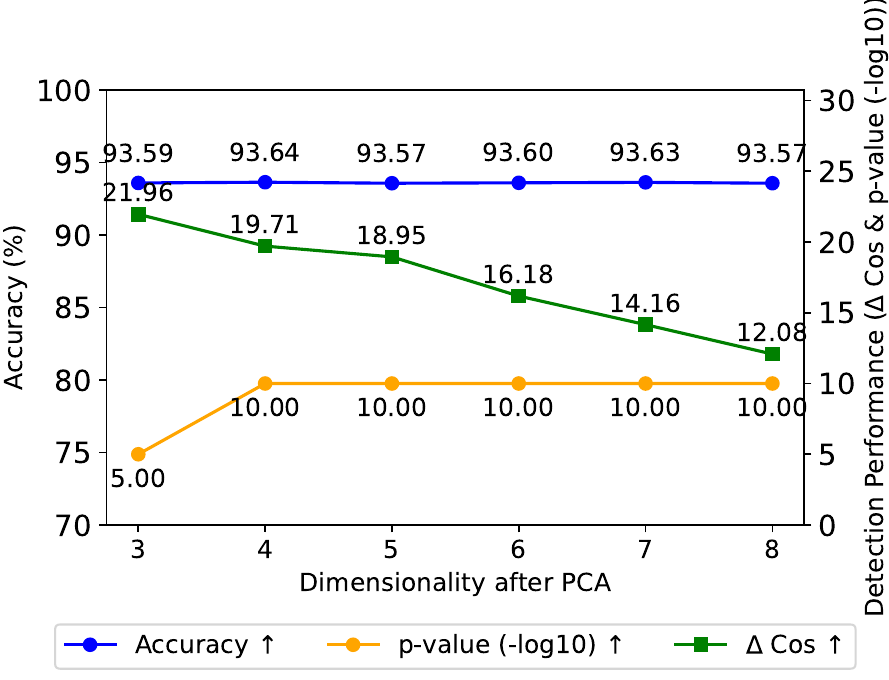}
        \caption{No Attack}
    \end{subfigure}
    \hfill
    \begin{subfigure}[b]{0.32\textwidth}
        \includegraphics[width=\textwidth]{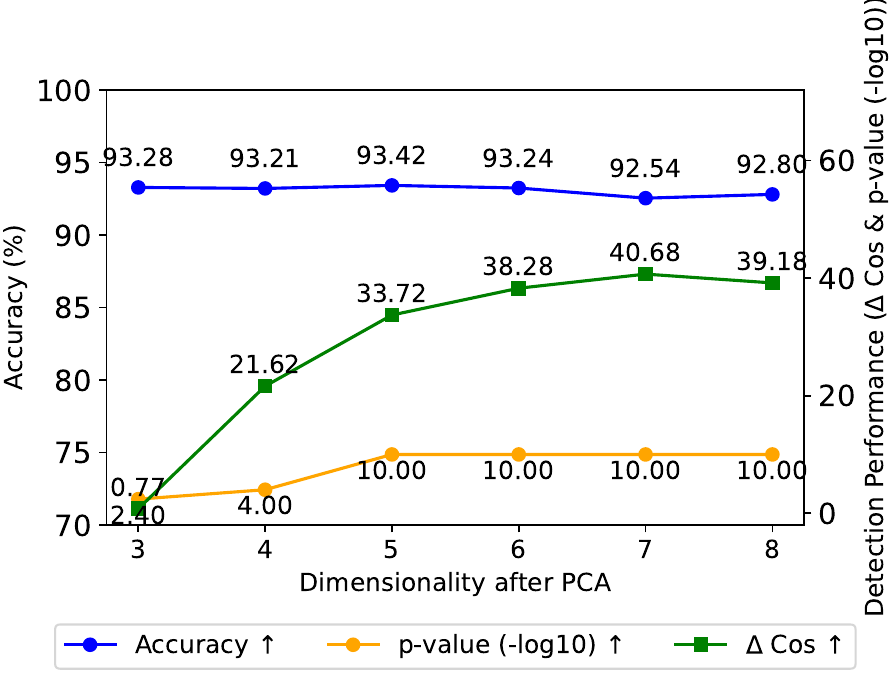}
        \caption{\ourattack}
    \end{subfigure}
    \hfill
    \begin{subfigure}[b]{0.32\textwidth}
        \includegraphics[width=\textwidth]{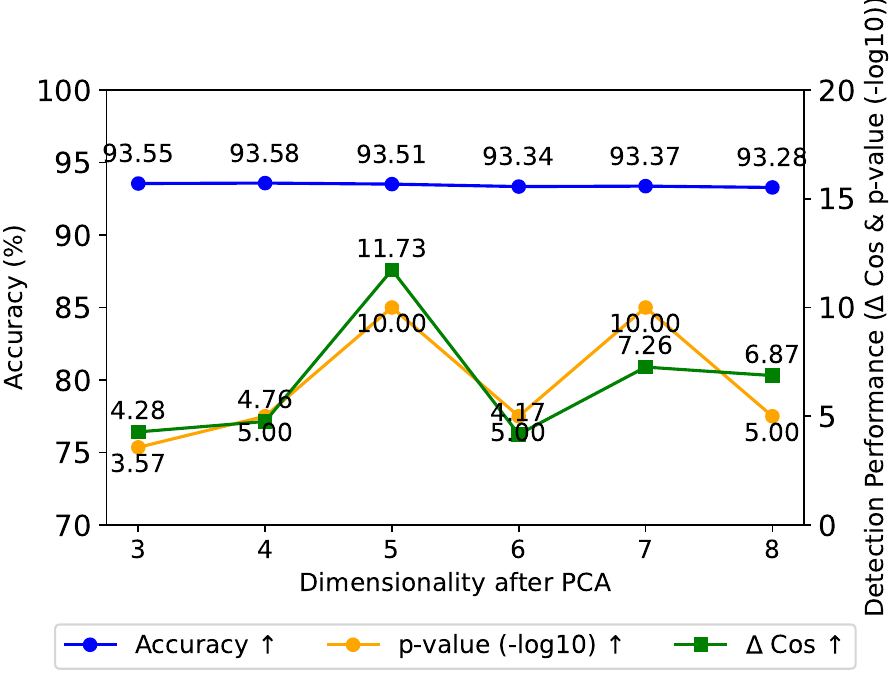}
        \caption{\dimshift}
    \end{subfigure}
    \caption{Impact of dimensionality after PCA under different attacks on the AG News dataset.}
    \label{fig:two_images_agw}
\end{figure*}

\begin{figure*}[t]
    \centering
    \begin{subfigure}[b]{0.32\textwidth}
        \includegraphics[width=\textwidth]{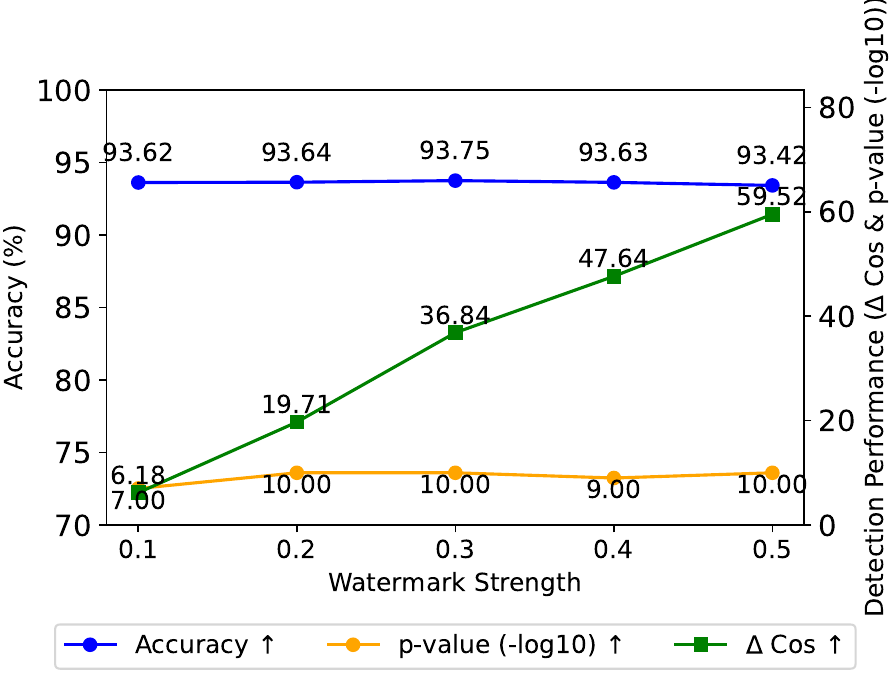}
        \caption{No Attack}
    \end{subfigure}
    \begin{subfigure}[b]{0.32\textwidth}
        \includegraphics[width=\textwidth]{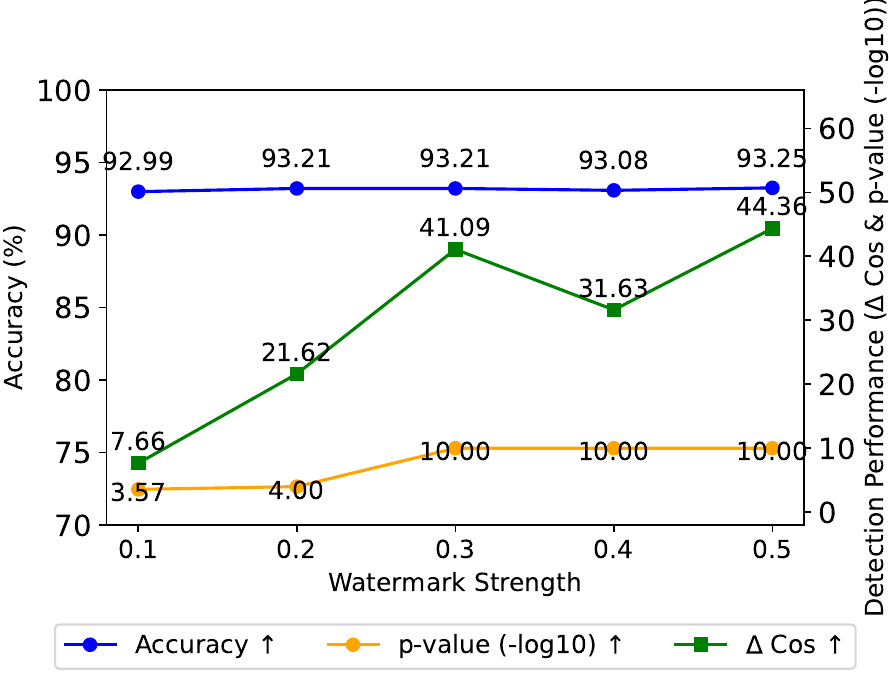}
        \caption{\ourattack}
    \end{subfigure}
    \begin{subfigure}[b]{0.32\textwidth}
        \includegraphics[width=\textwidth]{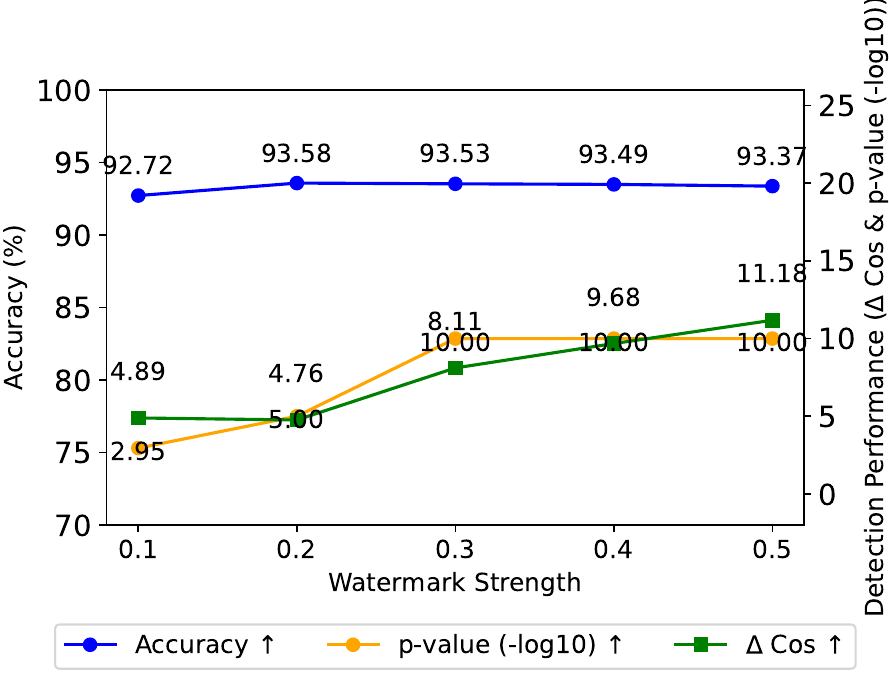}
        \caption{\dimshift}
    \end{subfigure}
    \caption{Impact of watermark strength $\lambda$ under different attacks on the AG News dataset.}
    \label{fig:watermark_weight_ag}
\end{figure*}

In this subsection, we present the results of watermark region ratio $\alpha$, dimensionality after PCA, and watermark strength $\lambda$ on the SST2, MIND and AG News dataset, as shown in Figures~\ref{fig:two_images_sst2b}--\ref{fig:watermark_weight_ag}.

The results on the three datasets are generally consistent with those in Section 4.4.
As the watermark region ratio $\alpha$ increases, the detection performance consistently improves across different datasets and attack settings. 
Meanwhile, as the dimensionality after PCA increases, detection performance tends to decline in the absence of attacks, but shows a fluctuating upward trend under attack conditions. 
In addition, increasing the watermark strength $\lambda$ also has a positive impact on detection effectiveness.
We set the watermark strength $\lambda$ to 0.2, which has negligible impact on task performance and achieves a good balance between utility and detection accuracy.

\begin{table*}[t] \small
\centering
\setlength{\tabcolsep}{1pt}
    \begin{tabular}{lcccccc}
    \toprule
    \multirow{3}{*}{\textbf{Method}} & \multicolumn{2}{c}{\textbf{Task Performance}} & \multicolumn{3}{c}{\textbf{Detection Performance}} & \multirow{3}{*}{\textbf{COPY?}} \\
    \cmidrule(lr){2-3} \cmidrule(lr){4-6}
    {} & \textbf{ACC.(\%)} & \textbf{$F_1$-score\xspace} & \textbf{{$p$-value $\downarrow$}} & \textbf{{$\Delta_{cos}(\%) \uparrow$}} & \textbf{{$\Delta_{l2}(\%) \downarrow$}}\\
    \midrule
    \swd + \gpt & 92.35$\pm$0.11 & 92.35$\pm$0.11 & {$< 10^{-5}$} & 7.35$\pm$2.21 & -14.70$\pm$4.41 & \cmark\\
    $\swd_{w/o PCA}$ + \gpt & 92.48$\pm$0.06 & 92.48$\pm$0.06 & {$< 0.01$} & 3.85$\pm$1.85 & -7.71$\pm$3.70 & \cmark\\
    $\swd_{single\ watermark}$ + \gpt & 92.57$\pm$0.15 & 92.56$\pm$0.15 & {$< 10^{-4}$} & 11.15$\pm$3.39 & -22.29$\pm$6.78 & \cmark\\
    \midrule
    \swd + \dimshift & 93.73$\pm$0.14 & 93.73$\pm$0.14 & $< 0.003$ & 2.77$\pm$0.47 &  -5.55$\pm$0.94 & \cmark\\
    $\swd_{w/o PCA}$ + \dimshift & 93.28$\pm$0.40 & 93.28$\pm$0.4 & {$< 0.05$} & 1.25$\pm$0.09 & -2.50$\pm$0.19 & \cmark\\
    $\swd_{single\ watermark}$ + \dimshift & 93.52$\pm$0.32 & 93.52$\pm$0.32 & {$< 10^{-3}$} & 2.40$\pm$0.71 & -4.80$\pm$1.43 & \cmark\\
    \bottomrule
    \end{tabular}
    \caption{Ablation study of our proposed method under two additional attacks on the SST-2 dataset.} 
    \label{table:more-necessity-pca}
\end{table*}

\subsection{Ablation Study Under Two Additional Attacks}
We further conducted an ablation study under paraphrasing and dimensional perturbation attacks.
As shown in Table \ref{table:more-necessity-pca}, in most cases, removing the dimensionality reduction component or using a single watermark results in a slight decline in detection performance.
These findings further highlight the importance of incorporating dimensionality reduction and multiple watermark embeddings in RegionMarker.

\subsection{Visualization of RegionMarker}
We visualize the embeddings before and after watermarking using t-SNE in Figure \ref{fig:visualization}.
We can observe that the embeddings of watermarked texts from a trigger region still cluster together, as the watermarked texts within that region share the same watermark embedding.
This makes it difficult for attack methods based on local structures to identify the backdoor texts.
In particular, because the embeddings of paraphrased texts remain within the trigger region, the watermark is preserved, making paraphrasing attacks ineffective.
Moreover, since the dimensionality reduction matrix is also unknown, it makes it very difficult for attackers to identify the watermarked texts.

\begin{table*}[h] \small

\centering
    \begin{tabular}{cccccccc}
    \toprule
    \multirow{2}{*}{Dataset} & \multirow{2}{*}{Method} & \multicolumn{2}{c}{Task Performance} & \multicolumn{3}{c}{Detection Performance} & \multirow{3}{*}{COPY?} \\
    \cmidrule(lr){3-4} \cmidrule(lr){5-7}
    {} & {} & ACC.(\%) & $F_1$-score\xspace & {$p$-value $\downarrow$} & {$\Delta_{cos}(\%) \uparrow$} & {$\Delta_{l2}(\%) \downarrow$}\\
    \midrule
    \multirow{4}{*}{\sst} & PCA & 93.23$\pm$0.36 & 93.23$\pm$0.36 & {$< 10^{-4}$} & 11.90$\pm$3.75 & -23.80$\pm$7.50 & \cmark \\
    & UMAP & 93.23$\pm$0.28 & 93.23$\pm$0.28 & - & - & - & \xmark \\
    & SVD & 93.21$\pm$0.28 & 93.21$\pm$0.29 & {$< 10^{-6}$} & 11.14$\pm$3.23 & -22.29$\pm$6.46 & \cmark \\
    & AE & 93.67$\pm$0.15 & 93.67$\pm$0.15 & {$< 10^{-4}$} & 11.28$\pm$1.18 & -22.56$\pm$2.36 & \cmark \\
    \midrule
    \multirow{4}{*}{\agnews} &PCA & 93.64$\pm$0.09 & 93.63$\pm$0.09 & $< 10^{-10}$ & 19.71$\pm$2.36 & -39.42$\pm$4.72 &\cmark \\
     &UMAP & 93.68$\pm$0.09 & 93.68$\pm$0.09 & $< 10^{-2}$ & 15.23$\pm$11.56 & -30.46$\pm$23.12 &\cmark \\
     &SVD & 93.61$\pm$0.10 & 93.60$\pm$0.10 & {$<10^{-9} $} & 19.83$\pm$2.14 & -39.65$\pm$4.28 &\cmark \\
     &AE & 93.62$\pm$0.12 & 93.61$\pm$0.12 & {$<10^{-9} $} & 21.56$\pm$0.89 & -43.12$\pm$1.77 &\cmark \\
    \bottomrule
    \end{tabular}
    \caption{The performance of four dimensionality reduction methods on two datasets.} 
    \label{table:different-dimensionality}
\end{table*}

\subsection{Effects of Different Dimensionality Reduction Methods} \label{app:dr}
We further explore the effects of different dimensionality reduction methods on two datasets, including PCA, UMAP, SVD, and Auto-Encoder. 

As shown in Table \ref{table:different-dimensionality}, UMAP is the least effective, the data distribution after UMAP is not uniform enough, and there is even no data projected to the watermarked region on the SST2 dataset.
This highlights the importance of uniformity for the effectiveness of semantic space partitioning.
The remaining three methods can achieve similarly satisfactory defense results.

\begin{table*}[t]
    \centering
    \small

    \begin{tabular}{ccccccc}
    \toprule
    \multirow{2}{*}{Dataset} & \multirow{2}{*}{Model} & \multirow{2}{*}{Parameters} & \multicolumn{3}{c}{Detection Performance} & \multirow{3}{*}{COPY?}\\
    \cmidrule(lr){4-6}
    {} & {} & {} & {$p$-value $\downarrow$} & {$\Delta_{cos}(\%) \uparrow$} & {$\Delta_{l2}(\%) \downarrow$} \\
    \midrule
    \multirow{5}{*}{SST2} & BERT-small     & 29M   & {$< 10^{-4}$} & 9.55  & -19.09 & \cmark \\
                          & BERT-base      & 108M  & {$< 10^{-4}$} & 11.90 & -23.80 & \cmark \\
                          & BERT-large     & 333M  & {$< 10^{-5}$} & 12.97 & -25.93 & \cmark \\
                          & Qwen2-0.5b     & 0.5B  & {$< 10^{-5}$} & 12.57 & -25.15 & \cmark \\
                          & Qwen2-1.5b     & 1.5B  & {$< 10^{-4}$} & 9.95  & -19.82 & \cmark \\
    \midrule
    \multirow{5}{*}{AG News} & BERT-small     & 29M   & {$< 10^{-10}$} & 18.91 & -37.82 & \cmark \\
                            & BERT-base      & 108M  & {$< 10^{-10}$} & 19.71 & -39.42 & \cmark \\
                            & BERT-large     & 333M  & {$< 10^{-10}$} & 19.71 & -39.43 & \cmark \\
                            & Qwen2-0.5b     & 0.5B  & {$< 10^{-7}$}  & 19.94 & -39.88 & \cmark \\
                            & Qwen2-1.5b     & 1.5B  & {$< 10^{-10}$} & 18.68 & -37.32 & \cmark \\
    \midrule
    \multirow{5}{*}{Enron} & BERT-small     & 29M   & {$< 10^{-4}$} & 10.80 & -21.60 & \cmark \\
                           & BERT-base      & 108M  & {$< 10^{-5}$} & 11.91 & -23.81 & \cmark \\
                           & BERT-large     & 333M  & {$< 10^{-5}$} & 11.93 & -23.85 & \cmark \\
                           & Qwen2-0.5b     & 0.5B  & {$< 10^{-8}$} & 16.86 & -33.73 & \cmark \\
                           & Qwen2-1.5b     & 1.5B  & {$< 10^{-4}$} & 12.00 & -23.99 & \cmark \\
    \midrule
    \multirow{5}{*}{MIND} & BERT-small     & 29M   & {$< 10^{-5}$}  & 14.11 & -28.23 & \cmark \\
                          & BERT-base      & 108M  & {$< 10^{-10}$} & 15.67 & -31.34 & \cmark \\
                          & BERT-large     & 333M  & {$< 10^{-4}$}  & 16.48 & -32.96 & \cmark \\
                          & Qwen2-0.5b     & 0.5B  & {$< 10^{-6}$}  & 16.14 & -32.27 & \cmark \\
                          & Qwen2-1.5b     & 1.5B  & {$< 10^{-7}$}  & 14.84 & -29.64 & \cmark \\
    
    \bottomrule
    \end{tabular}
    \caption{The impact of different extraction models on the different datasets.}
    \label{table:size}
\end{table*}

\begin{table*}[t]
\centering
\small
\setlength{\tabcolsep}{8pt}
\begin{tabular}{lcc}
\toprule
\textbf{Method} & \textbf{Injection Time (s/sample)} & \textbf{Detection Time (s)} \\
\midrule
WARDEN  & $2.11 \times 10^{-6}$  & 0.1440 \\
WET    & $1.09 \times 10^{-4}$   & 7.4499 \\
EspeW  & $3.11 \times 10^{-5}$   & 2.1231 \\
\textbf{RegionMarker (Ours)}   & $5.57 \times 10^{-5}$ & 3.8027 \\
\bottomrule
\end{tabular}
\caption{Average per-sample watermark injection and detection time of different methods on the SST-2 dataset.}
\label{tab:time-overhead}
\end{table*}

\subsection{Effects of Different Extracted Models} \label{app:backbone}
To evaluate the effects of different extraction models on our defense results, we further explore three sizes of BERT \citep{DBLP:conf/naacl/DevlinCLT19} and two sizes of Qwen2 \citep{Qwen2} on four datasets.

As shown in Table \ref{table:size}, we observe that our defense is effective against extraction models of different architectures (i.e., encoder-only architecture and decoder-only architecture) and sizes (ranging from 29M to 1.5B).
This demonstrates the generalization of our defense method.

\subsection{Effects of Watermarking on Embeddings} 
To evaluate the effects of watermarking on the original embeddings, we analyze the distribution of cosine similarity between embeddings before and after watermarking. 

\begin{figure*}[t]
    \centering
    \begin{subfigure}[b]{0.45\textwidth}
        \includegraphics[width=\textwidth]{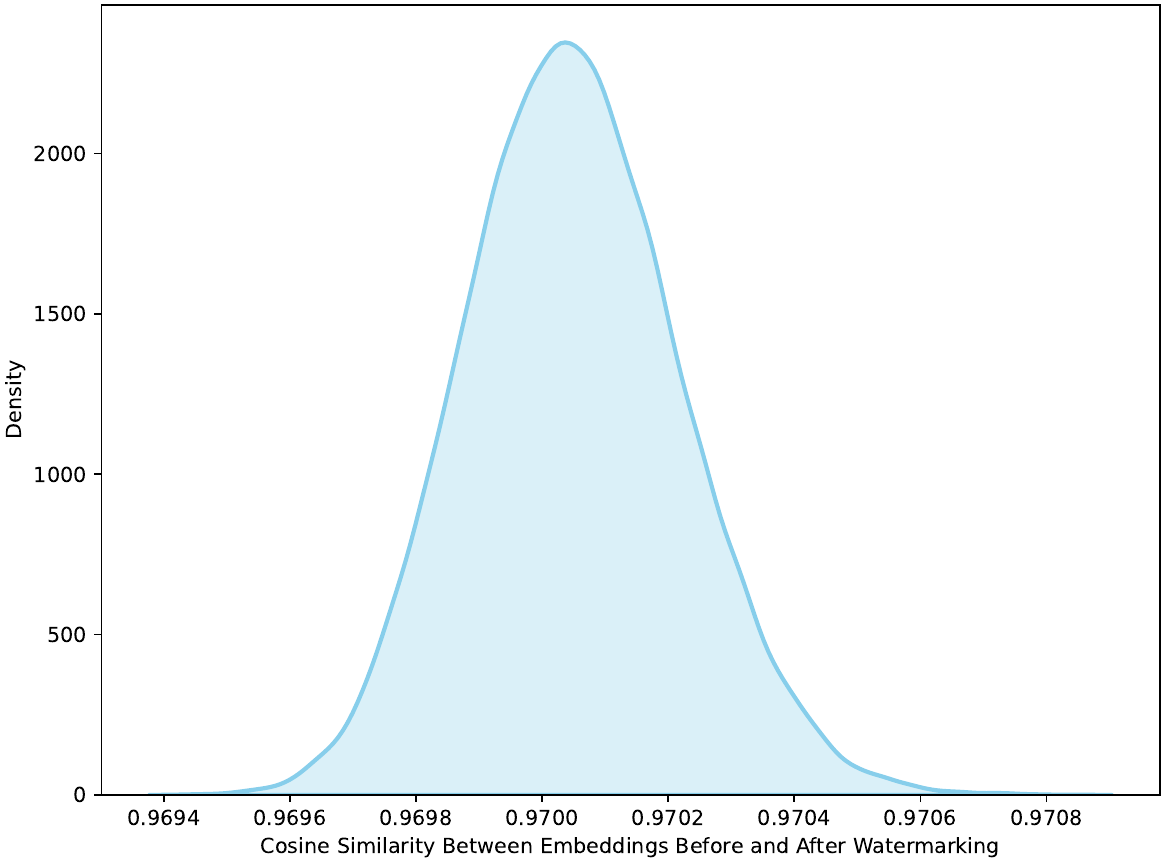}
        \caption{SST2}
    \end{subfigure}
    \begin{subfigure}[b]{0.45\textwidth}
        \includegraphics[width=\textwidth]{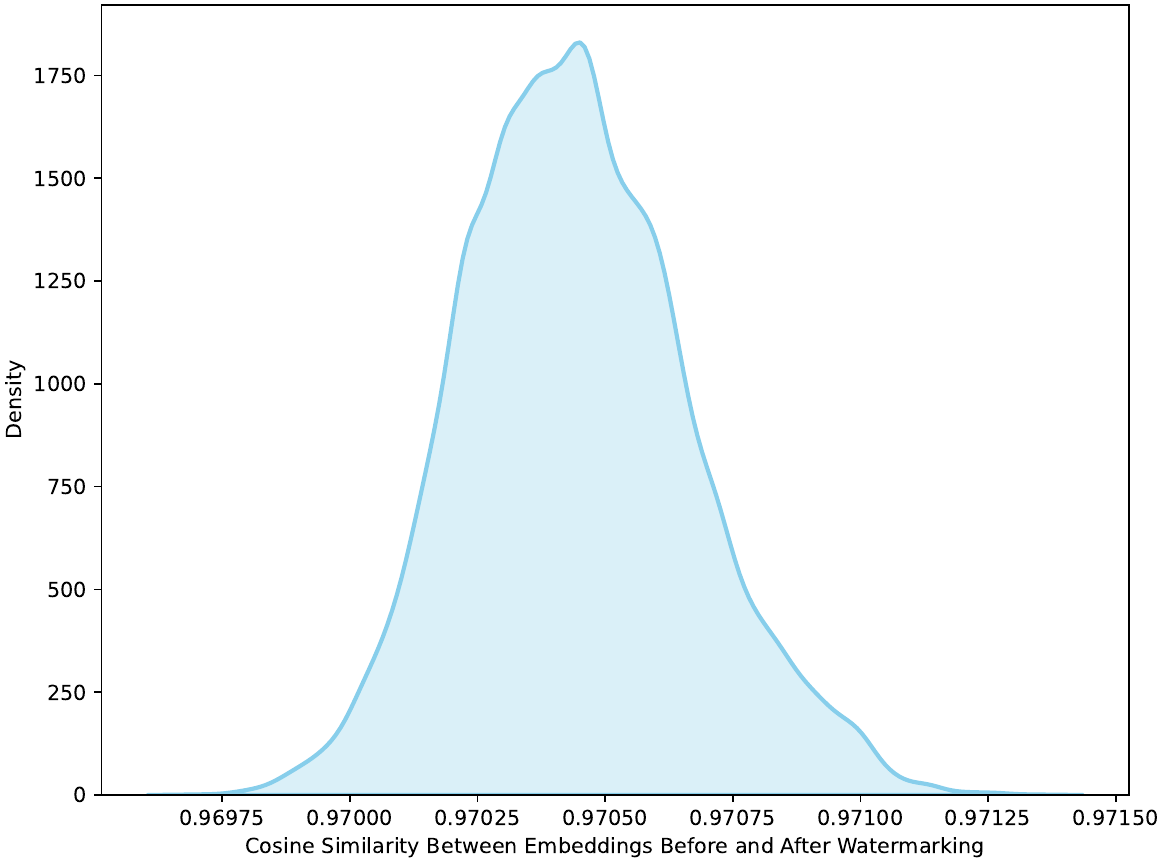}
        \caption{Enron}
    \end{subfigure}
    \begin{subfigure}[b]{0.45\textwidth}
        \includegraphics[width=\textwidth]{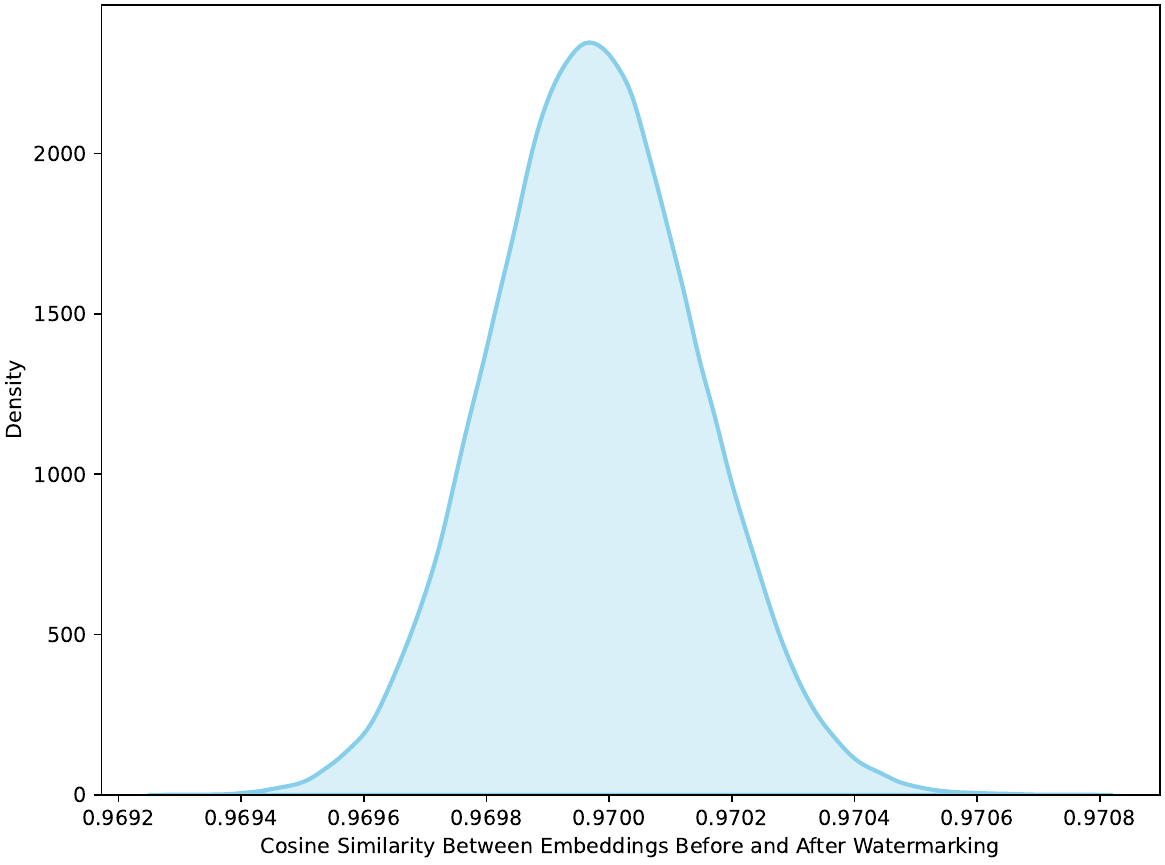}
        \caption{AG News}
    \end{subfigure}
    \begin{subfigure}[b]{0.45\textwidth}
        \includegraphics[width=\textwidth]{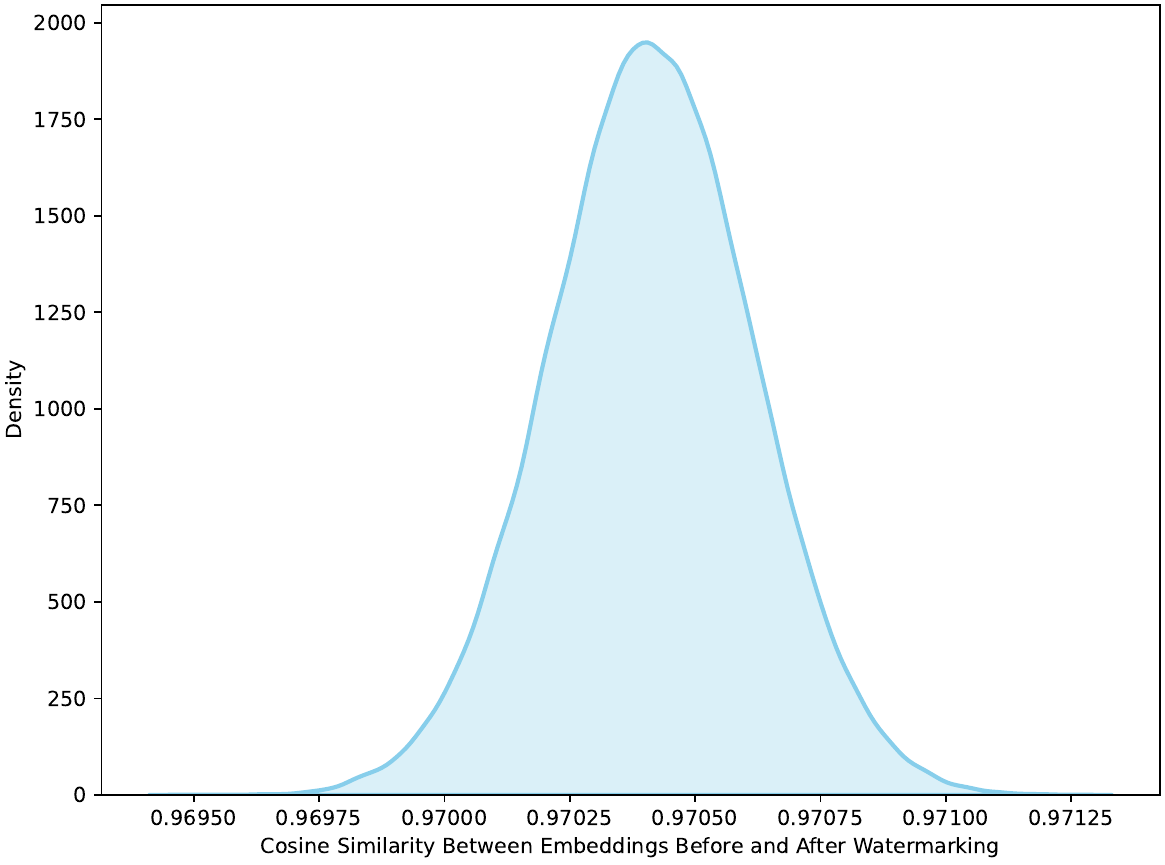}
        \caption{MIND}
    \end{subfigure}
    \caption{Distribution of Embedding Similarity Before and After Watermarking}
    \label{fig:cos_sim}
\end{figure*}

As shown in Figure \ref{fig:cos_sim}, across all four datasets, the similarity remains above 0.95, indicating that the watermark introduces only a negligible change to the original embeddings. 
This observation is further supported by Table~\ref{table:sst2-attack-performance},\ref{table:enron-attack-performance},\ref{table:mind-attack-performance}, and \ref{table:ag-attack-performance}, where the watermarking process has almost no effect on task performance. 
This is attributed to the small watermark strength (0.2), which ensures that the original embeddings are not significantly altered. 
These results demonstrate the stealthiness of our defense method.

\subsection{Time Overhead of Watermark Injection and Detection}
We evaluate the time overhead of watermark injection and detection across different methods.
As shown in Table~\ref{tab:time-overhead}, our method achieves an average injection time of less than 0.1 ms
ms per sample, which is negligible compared to the cost of generating embeddings.
Watermark detection also takes only a few seconds, making it highly efficient and practical for real-world use.
The injection and detection overheads of baseline methods are similarly low.
Overall, the additional time cost introduced by our method is minimal for model providers.

\subsection{Watermark Embedding and Detection Algorithms}
The region-triggered semantic watermark embedding and detection algorithms are shown in Algorithm~\ref{alg:embed} and Algorithm~\ref{alg:detect}, respectively.

\begin{algorithm}[t]
\caption{Region-Triggered Semantic Watermark Embedding Algorithm}
\label{alg:embed}
\textbf{Input}: Original embedding $e_0$, dimensionality reduction matrix $P$ \\
\textbf{Parameter}: Reduced dimension $d$, watermark strength $\lambda$, number of regions $2^d$ , watermark ratio $\alpha$\\
\textbf{Output}: Watermarked embedding $e_p$ \\
\begin{algorithmic}[1]
\STATE \textbf{Dimensionality Reduction:} \\
$v \leftarrow P e_0$ \ (project to $d$-dim semantic space)
\STATE \textbf{Region Partition (LSH)}: \\
For $i=1,\dots,d$, compute
\[
\text{LSH}_i(\mathbf{v})= \mathds{1}(\mathbf{n}_i \cdot \mathbf{v} > 0),
\]
and obtain the $d$-bit region signature $\text{LSH}(v) = [\text{LSH}_1(v),\dots,\text{LSH}_d(v)]$.
\STATE \textbf{Trigger Region Sampling}:\\
Randomly sample $R=\alpha \cdot 2^d$ regions as watermark regions $A=\{a_1,\dots,a_R\}$.
\STATE \textbf{Assign Watermarks}:\\
For each $a_r \in A$, assign a unique watermark embedding $w_r$ (e.g., the embedding of a target sample).
\IF{$\text{LSH}(v)$ corresponds to a trigger region $a_r \in A$}
    \STATE $\mathbf{e}_p \leftarrow  \text{Norm} ( (1-\lambda) \cdot \mathbf{e}_0 + \lambda \cdot \mathbf{w}_r )$
\ELSE
    \STATE $\mathbf{e}_p \leftarrow \mathbf{e}_0$
\ENDIF
\STATE \textbf{return} $\mathbf{e}_p$
\end{algorithmic}
\end{algorithm}

\begin{algorithm}[tb]
\caption{Region-Triggered Semantic Watermark Detection Algorithm}
\label{alg:detect}
\textbf{Input}: Watermark regions $A=\{a_1,\dots,a_R\}$, watermark set $W=\{w_1,\dots,w_R\}$, 
benign corpus $D_p^n$, backdoor corpus $\{D_{p}^{br}\}_{r=1}^R$ \\
\textbf{Output}: Infringement decision (is the target model extracted?)
\begin{algorithmic}[1]
\FOR{$r = 1$ to $R$}
    \STATE \textbf{Compute similarities/distances:}\\
    for every embedding $\mathbf{e}_i$:
    \[
    \text{cos}_{ir} =  \frac{\mathbf{e}_i \cdot \mathbf{w}_r}{||\mathbf{e}_i|| \cdot ||\mathbf{w}_r||}, l_{2ir} = { \left\| \frac{\mathbf{e}_i}{||\mathbf{e}_i||} - \frac{\mathbf{w}_r}{||\mathbf{w}_r||}  \right\| }^2
    \]
    \STATE $C_{b_r} \leftarrow \{\cos_{ir} \mid i \in D_{p}^{b_r}\}$ \\
    $C_{n_r} \leftarrow \{\cos_{ir} \mid i \in D_p^n\}$
    \STATE $L_{b_r} \leftarrow \{l_{2_{ir}} \mid i \in D_{p}^{b_r}\}$ \\
    $L_{n_r} \leftarrow \{l_{2_{ir}} \mid i \in D_p^n\}$
    \STATE \textbf{Compute differences}:
    \[
    \Delta_{cos_r} = \frac{1}{|C_{b_r}|}\sum_{i\in C_{b_r}} i -
    \frac{1}{|C_{n_r}|}\sum_{j\in C_{n_r}} j,
    \]
    \[
    \Delta_{l2_r} = \frac{1}{|L_{b_r}|}\sum_{i\in L_{b_r}} i -
    \frac{1}{|L_{n_r}|}\sum_{j\in L_{n_r}} j
    \]
    \STATE Compute $p\text{-value}_r$ (distribution difference between $C_{b_r}$ and $C_{n_r}$)
\ENDFOR
\STATE \textbf{Conservative aggregation}:
\[
\Delta_{\cos} = \max_{1\le r \le R} \Delta_{cos_r},
\]
\[
\Delta_{l2} = \min_{1\le r \le R} \Delta_{l2_r},
\]
\[
p\text{-value} = \min_{1\le r \le R} p\text{-value}_r
\]
\STATE If $p\text{-value} < \tau_p$ (e.g., 0.05) or $\Delta_{\cos}$ and $\Delta_{l2}$ exceed the threshold, 
decide \textbf{copyright infringement (model extraction)}.
\STATE \textbf{return} infringement decision
\end{algorithmic}
\end{algorithm}

\end{document}